\documentclass{article}

\usepackage{arxiv}

\usepackage[utf8]{inputenc} 
\usepackage[T1]{fontenc}    
\usepackage{hyperref}       
\usepackage{url}            
\usepackage{booktabs}       
\usepackage{amsfonts}       
\usepackage{nicefrac}       
\usepackage{microtype}      

\usepackage{lipsum}         
\usepackage{graphicx}
\usepackage{doi}

\usepackage{amsmath}
\usepackage{amssymb}
\usepackage{bm} 
\usepackage{makecell} 
\usepackage{multirow} 
\usepackage{colortbl} 
\usepackage{xcolor}   
\usepackage[numbers,sort&compress]{natbib}
\usepackage{cleveref}

\title{Depth-Synergized Mamba Meets Memory Experts for All-Day Image Reflection Separation}

\newif\ifuniqueAffiliation
\uniqueAffiliationtrue

\ifuniqueAffiliation 
\author{ 
	{Siyan Fang} \\
	Huazhong University of Science and Technology\\
	\texttt{siyanfang@hust.edu.cn} \\
	\And
	{Long Peng} \\
	University of Science and Technology of China\\
	\texttt{longp2001@mail.ustc.edu.cn} \\
	\And
	{Yuntao Wang} \\
	Huazhong University of Science and Technology\\
	\texttt{yuntaowang@hust.edu.cn} \\
	\And
	{Ruonan Wei} \\
	Huazhong University of Science and Technology\\
	\texttt{ruonan2765@gmail.com} \\
	\And
	{Yuehuan Wang}\thanks{Corresponding author.} \\
	Huazhong University of Science and Technology\\
	\texttt{yuehwang@hust.edu.cn} \\
}
\else
\fi

\date{}

\begin{document}
\maketitle

	\begin{abstract}
	Image reflection separation aims to disentangle the transmission layer and the reflection layer from a blended image. Existing methods rely on limited information from a single image, tending to confuse the two layers when their contrasts are similar, a challenge more severe at night. To address this issue, we propose the Depth-Memory Decoupling Network (DMDNet). It employs the Depth-Aware Scanning (DAScan) to guide Mamba toward salient structures, promoting information flow along semantic coherence to construct stable states. Working in synergy with DAScan, the Depth-Synergized State-Space Model (DS-SSM) modulates the sensitivity of state activations by depth, suppressing the spread of ambiguous features that interfere with layer disentanglement. Furthermore, we introduce the Memory Expert Compensation Module (MECM), leveraging cross-image historical knowledge to guide experts in providing layer-specific compensation. To address the lack of datasets for nighttime reflection separation, we construct the Nighttime Image Reflection Separation (NightIRS) dataset. Extensive experiments demonstrate that DMDNet outperforms state-of-the-art methods in both daytime and nighttime.
\end{abstract}

\noindent\textbf{Project Page:} \url{https://github.com/fashyon/DMDNet}


\section{Introduction}
Reflection artifacts often occur when capturing images through transparent media such as glass, not only compromising visual quality but also degrading the performance of downstream vision tasks \cite{kirillov2023segment,wang2024mfrgn,peng2024lightweight,10887773,pan2025enhance}. The task of image reflection separation aims to decompose a blended image $\boldsymbol{I}$ into a transmission layer $\boldsymbol{T}$ and a reflection layer $\boldsymbol{R}$, where $\boldsymbol{T}$ represents the scene behind the glass and $\boldsymbol{R}$ represents the reflected content on the glass surface. Early studies mainly rely on physical priors such as gradient sparsity \cite{levin2007user} and reflection blurriness \cite{fan2017generic,yang2019fast}, using handcrafted constraints based on physical assumptions. However, these methods are only effective in constrained scenarios.
\begin{figure}[t]
	\centering
	\includegraphics[width=0.63\columnwidth]{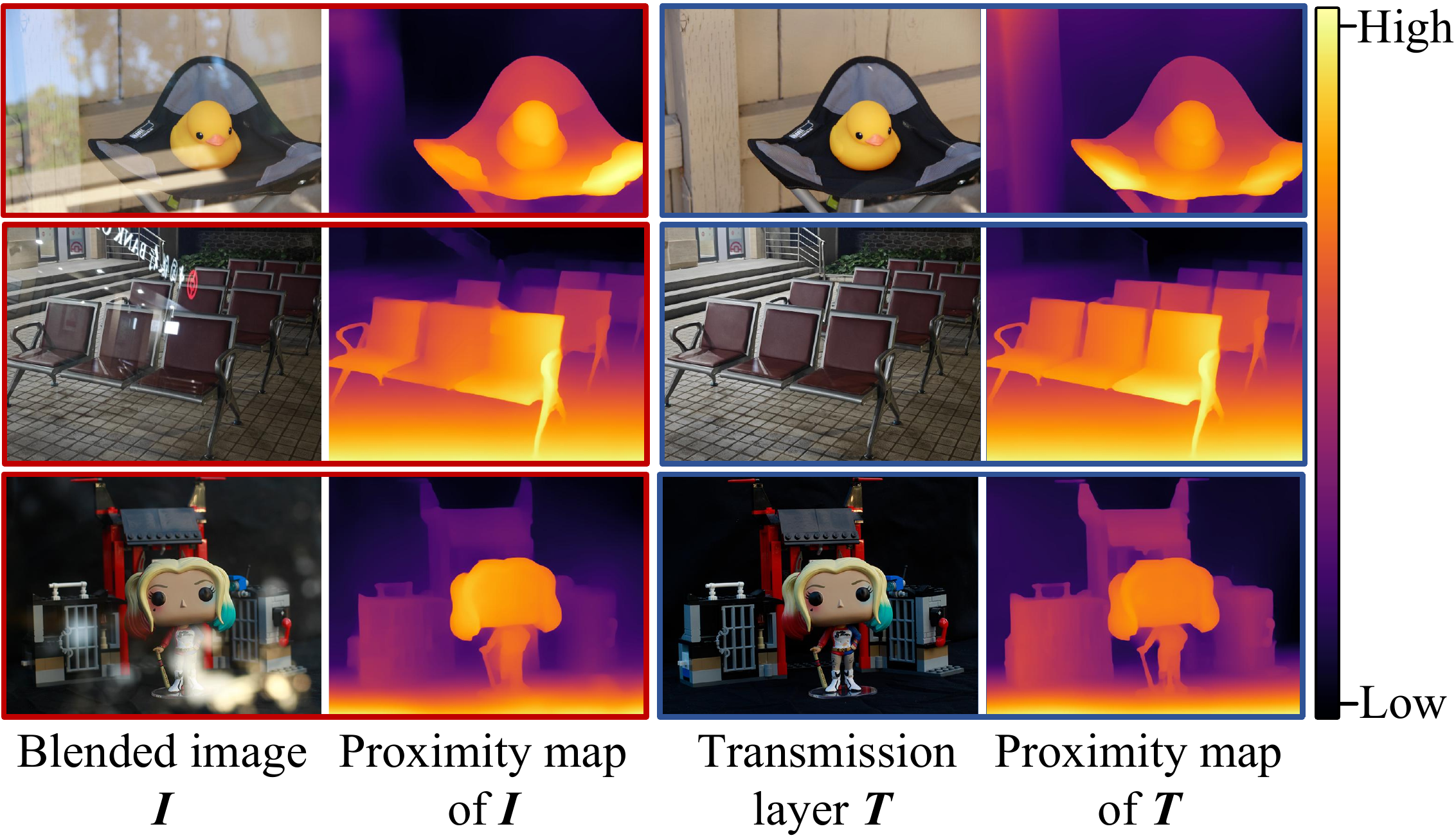} 
	\caption{Proximity maps obtained by depth estimation across daytime, nighttime, and indoor scenes. Depth estimation sees through reflection occlusion to capture the underlying structures of $\boldsymbol{T}$.}
	\label{fig_depth}
\end{figure}
With the development of deep learning~\cite{
	peng2020cumulative,
	peng2021ensemble,
	peng2025boosting,
	he2024latent,
	he2024dual,
	wang2023brightness,
	wang2023decoupling,
	yakovenko2025aim,
	jin2024mipi,
	li2023ntire,
	ren2024ninth,
	wang2025ntire,
	peng2024efficient,
	peng2024towards,
	peng2024unveiling,
	peng2025pixel,
	wu2025dropout,
	du2024fc3dnet,
	ignatov2025rgb,
	conde2024real,
	ren2024ultrapixel,
	wu2025robustgs,
	yan2025textual,
	sun2024beyond,
	suntext,
	feng2025pmq,
	zhang2025vividface,
	xu2025camel,
	qi2025data,
	jiang2024dalpsr,
	pengboosting}, methods such as Zhang et al. \cite{zhang2018single} and DSIT \cite{hu2024single} learn implicit priors of $\boldsymbol{T}$ and $\boldsymbol{R}$ from data to achieve separation. However, due to the limited information in a single image, these methods often encounter bottlenecks when $\boldsymbol{T}$ and $\boldsymbol{R}$ exhibit similar contrast. 
The challenge becomes particularly severe in nighttime scenes. In the daytime, abundant natural illumination strengthens $\boldsymbol{T}$ while suppressing $\boldsymbol{R}$, resulting in a clear contrast between the two layers. At night, illumination comes from artificial light sources that are randomly distributed, leading to uneven lighting conditions. Consequently, $\boldsymbol{T}$ appears darker due to insufficient global illumination, while localized strong lights incident on the glass surface produce glare and scattered highlights. As a result, $\boldsymbol{T}$ and $\boldsymbol{R}$ exhibit similar contrast levels, making their separation more challenging.

Although these difficulties are not directly addressed, some studies attempt to compensate by introducing additional physical cues, such as multi-view images \cite{xue2015computational,niklaus2021learned}, polarizing filters \cite{li2020reflection,lei2020polarized}, infrared cameras \cite{sun2019multi,hong2022reflection}, and flash illumination \cite{lei2021robust,wang2025flash}. However, such methods require controlled environments and extra devices, limiting their flexibility in applications. To eliminate reliance on external hardware, some studies incorporate human interaction, such as language prompts \cite{zhong2024language,hong2024differ} and manual region annotation \cite{zhang2019fast,chen2025firm}. Nevertheless, these approaches are time-consuming and labor-intensive.

Depth estimation offers physical cues without additional hardware or manual intervention.
By performing depth estimation \cite{ranftl2022towards} on blended images, we observe that the resulting proximity map highlights coherent and sharp structures corresponding to $\boldsymbol{T}$, while blurry and transparent overlays associated with $\boldsymbol{R}$ are naturally suppressed, as shown in Figure~\ref{fig_depth}. This indicates that high proximity values tend to carry salient structures. These structures often span large spatial ranges, such as the outline of a building or a row of chairs, fully exploiting these cues requires a model capable of capturing long-range dependencies. 


Mamba \cite{gu2023mamba} has achieved impressive results in various fields \cite{zhu2024vision,peng2025directing}, enabled by the efficient long-range modeling of its State-Space Model (SSM). VMamba \cite{liu2024vmamba} brings this capability to the vision domain through four-directional scanning. However, this scanning strategy has two limitations for image reflection separation:

(1) Disruption of Structural Continuity.
The transmission scene is typically defined by coherent contours, shapes, and textures, such as the edges of windows or the curves of human faces. The fixed sequential scanning fragments this content, leading to distorted structural cues while hindering the perception of these semantic entities as a whole.

(2) Error Propagation.
In SSM, the state of earlier-scanned regions continuously influences subsequent ones. If ambiguous features are propagated first, their uncertainty spreads throughout the entire image, amplifying separation errors.

To address these issues, we propose the Depth-Synergized Decoupling Mamba (DSMamba). Its Depth-Aware Scanning Strategy (DA-Scan) customizes scanning strategies separately for $\boldsymbol{T}$ and $\boldsymbol{R}$, allowing the model to encounter salient structures at early stages of modeling, helping to establish semantic continuity. In synergy with DA-Scan, we design the Depth-Synergized State-Space Model (DS-SSM) to modulate the activity of state evolution while suppressing activations in ambiguous areas, preventing the spread of erroneous information.

To overcome the limited information of a single image, we introduce the Memory Expert Compensation Module (MECM) to leverage cross-image historical knowledge. Each expert is equipped with a memory bank that stores feature patterns, and MECM dynamically activates the most relevant experts to provide targeted compensation. For example, experts specialized in texture details and structural contours can be activated for $\boldsymbol{T}$, while those handling sparse highlights and blurred ghosting can be used for $\boldsymbol{R}$.

To address the scarcity of datasets for nighttime image reflection separation, we construct the Nighttime Image Reflection Separation (NightIRS) dataset. It comprises 1,000 image triplets obtained under nighttime reflection conditions. This dataset captures the unique complexities of nighttime imaging, including uneven illumination, strong artificial light sources, and diverse reflection artifacts, which are often overlooked in existing public datasets.

Overall, the contributions of this work are as follows:
\begin{itemize}
	\item We propose DSMamba, with DA-Scan and DS-SSM working in synergy to guide Mamba toward structural saliency and suppress erroneous propagation.
	
	\item We introduce MECM to leverage cross-image historical memory for targeted compensation.
	
	\item We construct the NightIRS dataset for evaluating nighttime reflection separation.
	
	\item Experimental results demonstrate that DMDNet outperforms State-of-the-Art Methods (SOTAs).
\end{itemize}

\section{Related Work}

\textbf{Image Reflection Separation.}
Early studies \cite{levin2007user,yang2019fast} rely on handcrafted priors, which only work in simple cases. Deep learning methods \cite{zhang2018single,hu2024single} learn mappings from contaminated to clean images using large-scale data, but often struggle with complex scenes due to limited information in a single image. To incorporate physical cues, some approaches leverage  multi-view images, polarization \cite{lei2020polarized}, flash \cite{lei2021robust}, or infrared cameras \cite{hong2022reflection}, but these require extra hardware, making them unsuitable for internet images. To avoid this, Zhong et al. \cite{zhong2024language} introduce language prompts, while FIRM \cite{chen2025firm} relies on manual region annotations. However, these methods need human intervention and thus limit automation. In contrast, depth estimation offers physical cues without external sensors. Elnenaey et al. \cite{elnenaey2024utilizing} coarsely quantize the depth map into four levels and concatenate it with the input image for guidance. DGR$^{2}$-Net \cite{he2025rethinking} applies global pooling on the depth map and then concatenates it with the input for binocular reflection removal. However, these methods lack fine-grained depth guidance, resulting in inadequate effectiveness. More importantly, they overlook the structural saliency embedded in depth maps for image reflection separation.

\textbf{Visual Mamba.}
Due to Mamba's strong performance in long-sequence modeling, it has recently been widely adopted in various vision tasks~\cite{zhu2024vision,liu2024vmamba,xia2024s3mamba,di2025qmambabsr,he2024multi}. MambaIR \cite{guo2024mambair} and VMambaIR \cite{shi2025vmambair} are among the earliest works to introduce the Mamba into the field of image restoration. Subsequently, MambaIRv2 \cite{guo2025mambairv2} proposes a semantics-guided neighborhood interaction mechanism to facilitate information transfer. TAMambaIR \cite{peng2025directing} introduces a multi-directional receptive field expansion scheme to enhance modeling capability. However, these methods lack dynamic state modeling strategies sensitive to geometric structures, limiting their ability to distinguish between layers in reflection separation.

\textbf{Mixture of Experts (MoE).}
MoE enables adaptive computation by employing multiple experts, and has been widely applied to image restoration tasks. MoCE-IR \cite{zamfir2025complexity} designs expert modules with varying computational complexity to match different degradation. FAME \cite{he2024frequency} adopts a frequency-adaptive MoE architecture, applying different dynamic processing strategies to low- and high-frequency components. However, these methods lack cross-image memory, limiting their ability to compensate for contaminated information within a single image.

\textbf{Memory-Augmented Methods.}
Several studies explore memory mechanisms for image restoration. For instance, Xu et al. \cite{xu2021texture} propose a texture memory that stores patch samples to guide texture synthesis. ER$^{2}$Net \cite{zou2024eyeglass} leverages a memory module to inpaint eyeglass reflection regions. However, the high computational cost restricts it to one-off usage, making it unsuitable for the deployment of multiple experts. Moreover, they are limited to either global matching or local modeling, without a unified mechanism to enable adaptive expert behavior.

\begin{figure}[t]
	\centering
	\includegraphics[width=0.6\columnwidth]{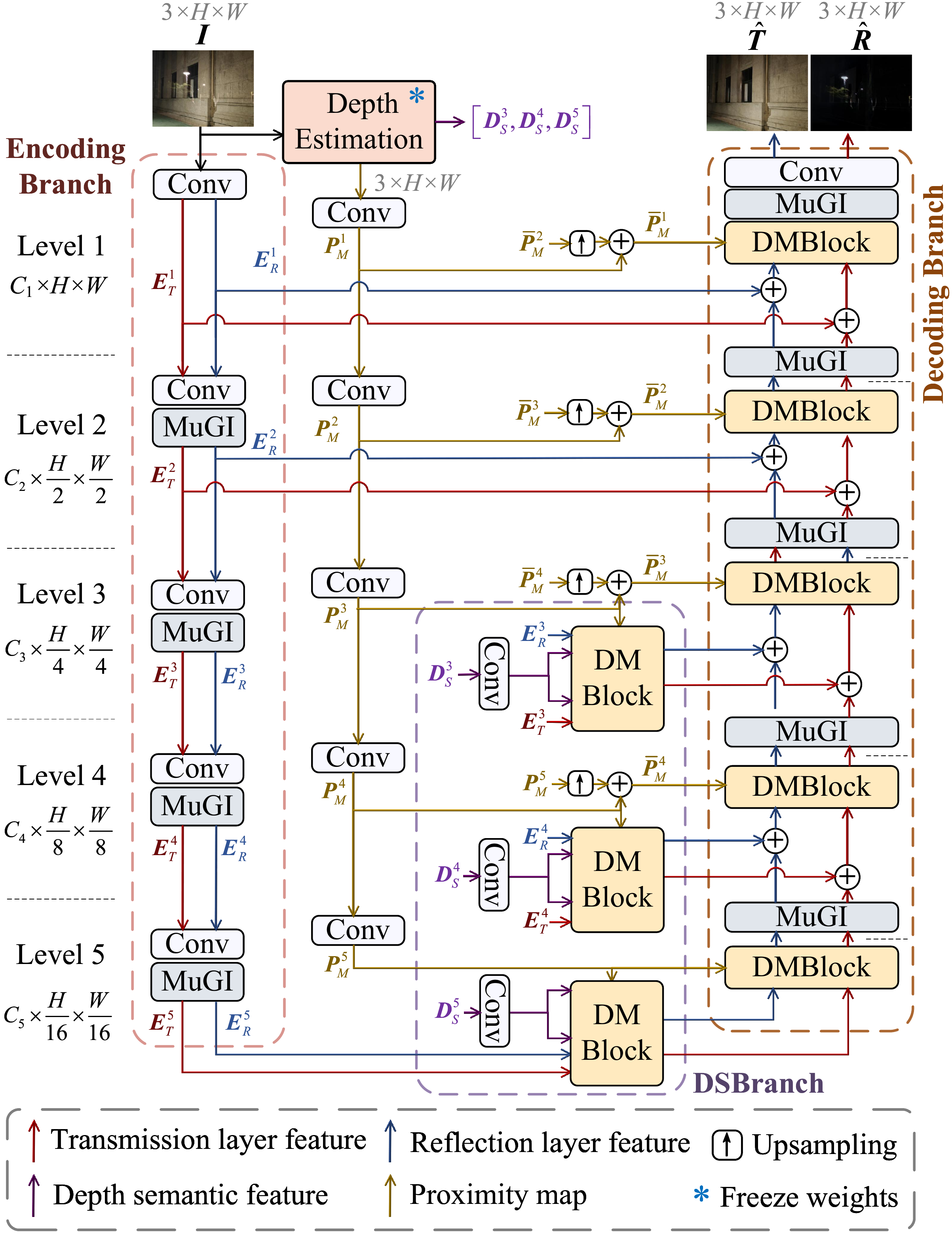} 
	\caption{Depth-Memory Decoupling Network (DMDNet). DMDNet employs the DMBlock to decouple $\boldsymbol{T}$ and $\boldsymbol{R}$ using depth and memory cues.}
	\label{fig_DMDNet}
\end{figure}


\begin{figure*}[t]
	\centering
	\includegraphics[width=0.99\textwidth]{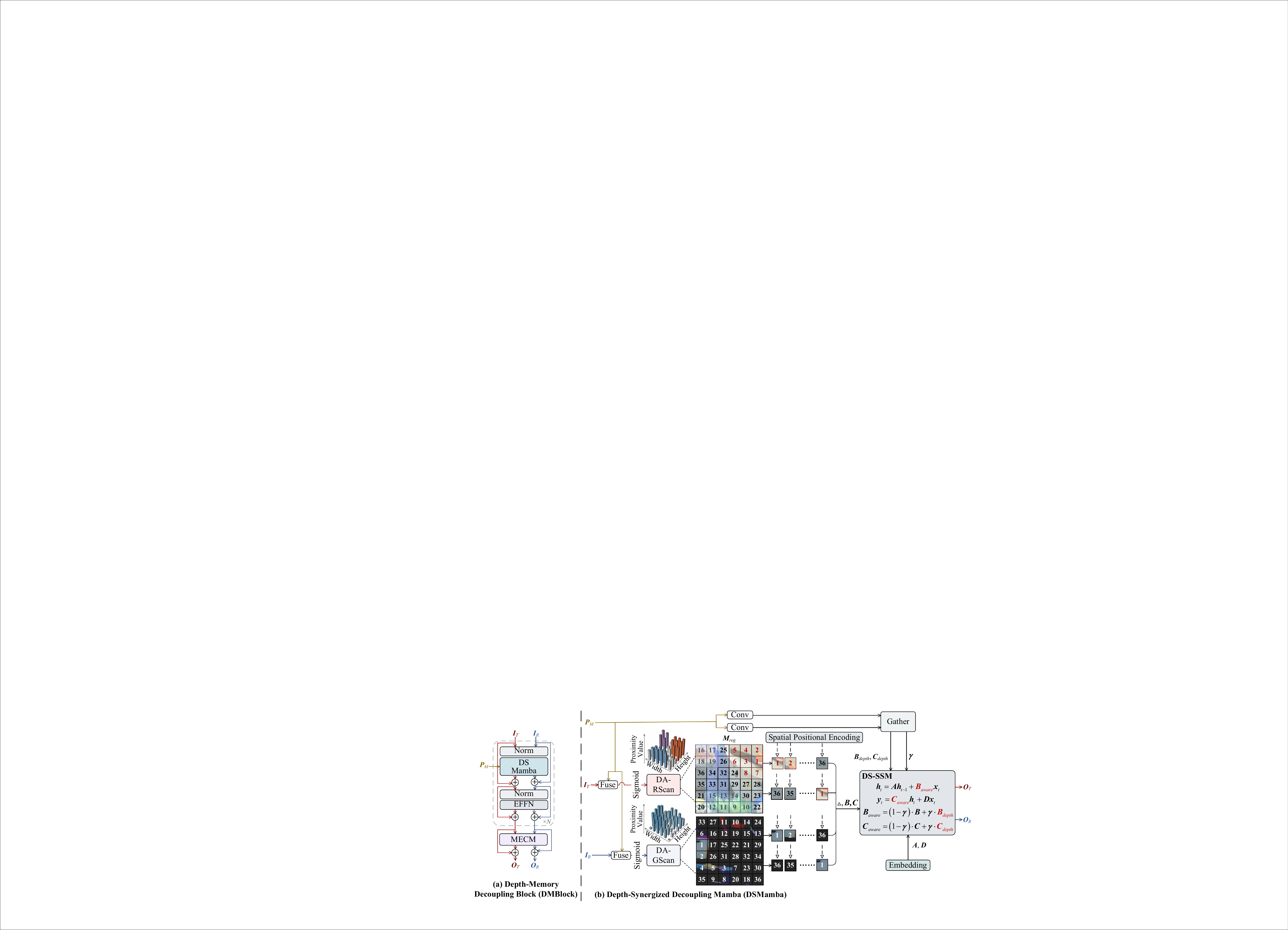} 
	\caption{DMBlock and DSMamba. DSMamba prioritizes salient structures via DAScan and synergistically modulates state activations through DS-SSM. The numbers indicate the forward scanning order.}
	
	\label{fig_DSMamba}
\end{figure*}

\section{Methodology}
\subsection{Depth-Memory Decoupling Network}
The Depth-Memory Decoupling Network (DMDNet) consists of the Encoding Branch, the Depth Semantic Modulation Branch (DSBranch), and the Decoding Branch, as shown in Figure~\ref{fig_DMDNet}.
The Encoding Branch adopts the Mutually-Gated Interactive Block (MuGI) \cite{hu2023single} to extract the features of $\boldsymbol{T}$ and $\boldsymbol{R}$, where
$\boldsymbol{E}^i_{T},\ \boldsymbol{E}^i_{R} \in \mathbb{R}^{C_i \times \frac{H}{2^{i-1}} \times \frac{W}{2^{i-1}}},\quad i \in \{1, 2, 3, 4, 5\}$. 
Here, $C_i$ denotes the number of channels at the $i$-th level, and $H$ and $W$ are the height and width of the input image $\boldsymbol{I}$, respectively.
The DSBranch leverages depth semantic features  
$\boldsymbol{D}^3_{S} \in \mathbb{R}^{96 \times \frac{H}{4} \times \frac{W}{4}},\quad
\boldsymbol{D}^4_{S} \in \mathbb{R}^{256 \times \frac{H}{8} \times \frac{W}{8}},\quad
\boldsymbol{D}^5_{S} \in \mathbb{R}^{512 \times \frac{H}{16} \times \frac{W}{16}}$, modulating the encoded features for the Decoding Branch.
The Decoding Branch performs the separation of $\boldsymbol{T}$ and $\boldsymbol{R}$ through the Depth-Memory Decoupling Block (DMBlock) and the proximity maps 
$\boldsymbol{P}^i_{M} \in \mathbb{R}^{C \times \frac{H}{2^{i-1}} \times \frac{W}{2^{i-1}}},\quad i \in \{1, 2, 3, 4, 5\}$.
As shown in Figure~\ref{fig_DSMamba}(a), the DMBlock consists of DSMamba, MECM, and EFFN \cite{shi2025vmambair}.

\subsection{Depth-Synergized Decoupling Mamba}
To address the limitation of Mamba’s fixed scanning strategy, we propose Depth-Synergized Decoupling Mamba (DSMamba). As illustrated in Figure \ref{fig_DSMamba}(b), DSMamba consists of the Depth-Aware Scanning (DAScan) and the Depth-Synergized State-Space Model (DS-SSM). 	The DAScan adopts Depth-Aware Regional Scanning (DA-RScan) for $\boldsymbol{T}$, and Depth-Aware Global Scanning (DA-GScan) for $\boldsymbol{R}$. 

DA-RScan follows a “large-area-first + near-to-far” scheme. Specifically, the proximity map is partitioned into a region scanning map $\boldsymbol{M}_{reg}$. Regions are scanned from the largest to the smallest, as larger regions indicate more salient semantics, with the background region scanned at the end to ensure completeness. This region-based scheme preserves the semantic continuity of pixels within the same object. Inside each region, pixels are scanned in a near-to-far order, prioritizing structurally salient structures. 

DA-GScan follows a “global near-to-far” scheme, scanning from the globally nearest pixels to the farthest. This scheme emphasizes global structural saliency, which matches the sparse and discontinuous distribution of $\boldsymbol{R}$ to enhance the modeling of reflection features. Finally, inverse DAScan is applied in the opposite order to complement structural cues.

The vanilla State Space Model (SSM) in Mamba adopts a uniform state update mechanism for all regions, formulated as:
\begin{equation}
	\boldsymbol{h}_t = \boldsymbol{A h}_{t-1} + \boldsymbol{B x}_t,\quad \boldsymbol{y}_t = \boldsymbol{C h}_t + \boldsymbol{D x}_t
\end{equation}
where $\boldsymbol{A} \in \mathbb{R}^{N \times N}$, $\boldsymbol{B} \in \mathbb{R}^{N \times 1}$, $\boldsymbol{C} \in \mathbb{R}^{1 \times N}$, and $\boldsymbol{D} \in \mathbb{R}$. $N$ is the state size.
This mechanism lacks structural awareness, making it difficult to disentangle regions where $\boldsymbol{T}$ and $\boldsymbol{R}$ are intricately intertwined. To overcome this constraint while synergizing with DAScan, we design the DS-SSM, whose state update is defined as:
\begin{equation}
	\begin{aligned}
		\boldsymbol{h}_t &= \boldsymbol{A} \boldsymbol{h}_{t-1} + \boldsymbol{B}_{\mathit{aware}} \boldsymbol{x}_t, \\
		\boldsymbol{y}_t &= \boldsymbol{C}_{\mathit{aware}} \boldsymbol{h}_t + \boldsymbol{D} \boldsymbol{x}_t, \\
		\boldsymbol{B}_{\mathit{aware}} &= (1 - \boldsymbol{\gamma}) \cdot \boldsymbol{B} + \boldsymbol{\gamma} \cdot \boldsymbol{B}_{\mathit{depth}}, \\
		\boldsymbol{C}_{\mathit{aware}} &= (1 - \boldsymbol{\gamma}) \cdot \boldsymbol{C} + \boldsymbol{\gamma} \cdot \boldsymbol{C}_{\mathit{depth}}
	\end{aligned}
\end{equation}
Here, $\boldsymbol{\gamma}$ is a weighting map between 0–1, derived from the proximity map. $\boldsymbol{B}_{\mathit{depth}}$ and $\boldsymbol{C}_{\mathit{depth}}$ are depth-guided state matrices that respectively control the magnitude of state updates and the contribution of the state to the output.

In structurally salient regions, a larger $\boldsymbol{\gamma}$ strengthens the influence of $\boldsymbol{B}_{{depth}}$ and $\boldsymbol{C}_{{depth}}$, accelerates the integration of clear structures, and reinforces their guidance on the output. Conversely, in structurally ambiguous regions, the intervention is suppressed to prevent the propagation of ambiguous features.


\textbf{Spatial Positional Encoding.}
To reinforce positional specificity during the scanning, DSMamba employs a Spatial Positional Encoding (SPE) based on 2D sine and cosine functions:
\begin{equation}
	\begin{aligned}
		\boldsymbol{PE}_x &= \left[\sin(x \cdot f_i),\ \cos(x \cdot f_i)\right], \\
		\boldsymbol{PE}_y &= \left[\sin(y \cdot f_i),\ \cos(y \cdot f_i)\right]
	\end{aligned}
\end{equation}
where $x$ and $y$ denote the normalized spatial coordinates, and $f_i$ represents different frequency bands.

By combining the horizontal and vertical encodings, a positional embedding $\boldsymbol{PE} \in \mathbb{R}^{H \times W \times d_{\mathit{inner}}}$ is obtained, where $d_{\mathit{inner}}$ is the channel dimension of the state-space model. The embedding is  realigned with the scanning order and added to the state features, providing positional cues for state modeling.
\begin{figure*}[t]
	\centering
	\includegraphics[width=0.99\textwidth]{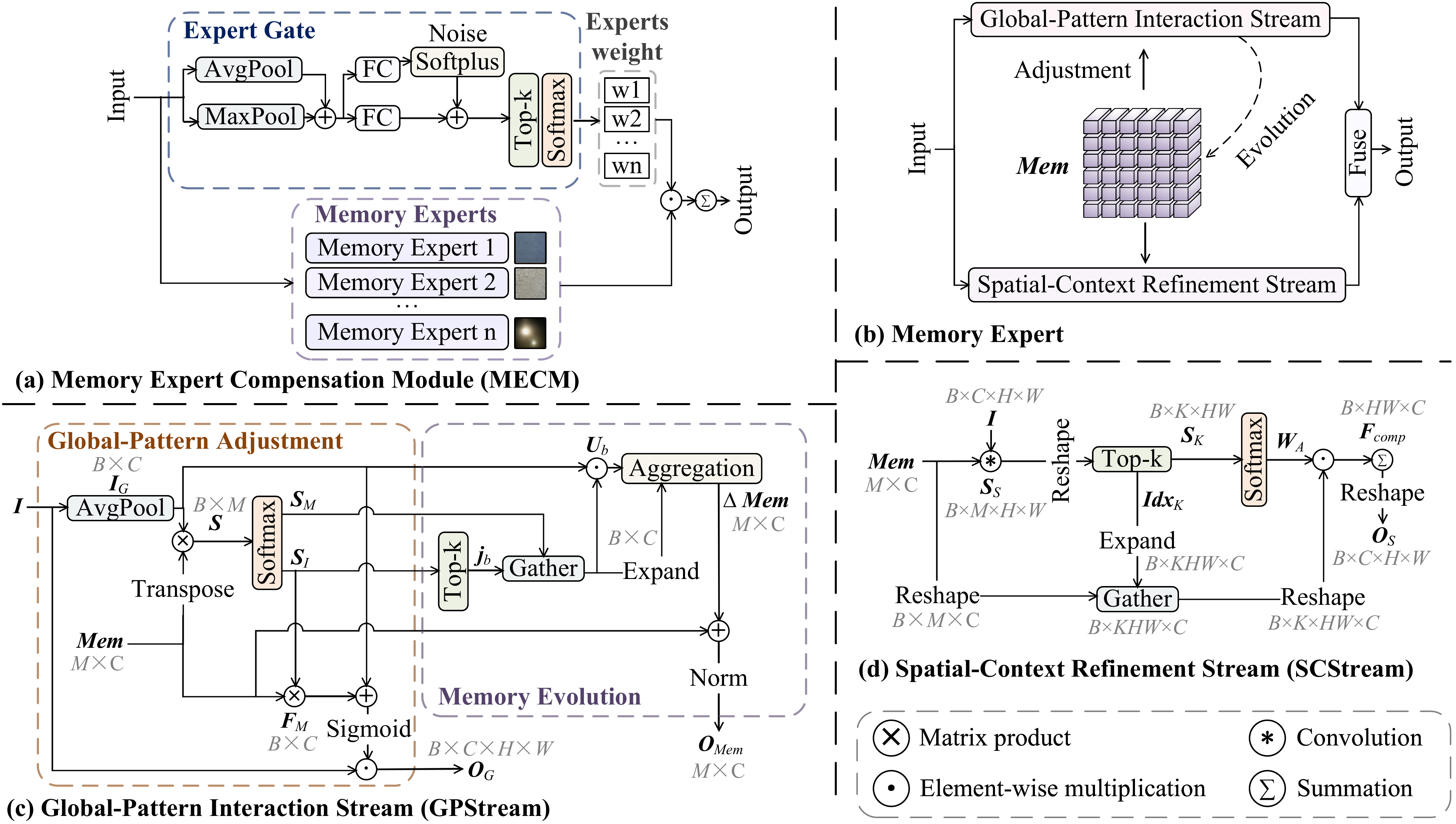} 
	\caption{Memory Expert Compensation Module (MECM) and its components, which leverage cross-image historical knowledge to guide the decoupling. Each memory expert consists of the GPStream for global adjustment and memory evolution, and the SCStream for spatial-level refinement.}	
	\label{fig_MECM}
\end{figure*}

\subsection{Memory Expert Compensation Module}
To leverage cross-image accumulated knowledge for targeted compensation, we introduce the Memory Expert Compensation Module (MECM), as illustrated in Figure \ref{fig_MECM}. MECM consists of the Expert Gate \cite{he2024frequency} and Memory Experts. The Expert Gate is responsible for selecting the most relevant $N_{\mathit{Exp}}^{K}$ experts from $N_{\mathit{Exp}}$ candidates. The Memory Experts perform feature retrieval and evolution, enabling adaptive compensation with historical knowledge.

The Memory Expert comprises the Global-Pattern Interaction Stream (GPStream) and the Spatial-Context Refinement Stream (SCStream).  The GPStream is further divided into Global-Pattern Adjustment and Memory Evolution.

For the Global-Pattern Adjustment, the input image $\boldsymbol{I} \in \mathbb{R}^{B \times C \times H \times W}$ is first pooled into a global representation $\boldsymbol{I}_{\mathit{G}} \in \mathbb{R}^{B \times C}$, which is used to compute similarity with the memory bank $\boldsymbol{Mem} \in \mathbb{R}^{M \times C}$, yielding a similarity score matrix $\boldsymbol{S} \in \mathbb{R}^{B \times M}$, where $B$ is the batch size and $M$ is the number of memory items. We apply softmax along the memory and image dimensions to obtain $\boldsymbol{S}_{\mathit{I}}$ and $\boldsymbol{S}_{\mathit{M}}$, respectively. Here, $\boldsymbol{S}_{\mathit{I}}$ denotes matching distribution of each image over all memory items, while $\boldsymbol{S}_{\mathit{M}}$ represents the contribution of each memory item to the image. Next, $\boldsymbol{S}_{{I}}$ is used to perform weighted aggregation of $\boldsymbol{Mem}$, producing the memory response feature $\boldsymbol{F}_{\mathit{M}} \in \mathbb{R}^{B \times C}$. Finally, $\boldsymbol{F}_{\mathit{M}}$ interacts with $\boldsymbol{I}_{\mathit{G}}$ to generate an attention mask that modulates the input $\boldsymbol{I}$, producing the global compensation $\boldsymbol{O}_{\mathit{G}}$.

Memory Evolution aims to provide feedback and update the memory bank.  
For each image sample $b \in [1, B]$, the most responsive memory index $j_b \in [1, M]$ is selected from the matching matrix $\boldsymbol{S}_{\mathit{I}}$. The corresponding score $\boldsymbol{S}_{\mathit{M}}[b, j_b]$ is used as a weight to perform multiplication with the global representation $\boldsymbol{I}_{\mathit{G}}[b]$, resulting in an update vector $\boldsymbol{U}_b \in \mathbb{R}^C$. All $\boldsymbol{U}_b$ vectors are  aggregated along their associated index $j_b$ to form a memory increment $\Delta \boldsymbol{Mem} \in \mathbb{R}^{M \times C}$:
\begin{equation}
	\Delta \boldsymbol{Mem}[m] = \sum_{b \in [1, B],\ j_b = m} \boldsymbol{U}_b, \quad m \in [1, M]
\end{equation}
Finally, the memory bank is updated in a residual manner to obtain the updated memory $\boldsymbol{O}_{\mathit{Mem}}$.

SCStream focuses on spatial contextual compensation. First, the memory bank $\boldsymbol{Mem}$ is reshaped as convolutional kernels and convolved with the input image $\boldsymbol{I}$ to obtain the similarity map $\boldsymbol{S}_{\mathit{S}} \in \mathbb{R}^{B \times M \times H \times W}$. $\boldsymbol{S}_{\mathit{S}}[b, m, h, w]$ denotes the similarity between location $(h, w)$ and the $m$-th memory item. Next, for each spatial position, the Top-k most relevant memory items are selected. Specifically, $\boldsymbol{Idx}_{\mathit{K}},\ \boldsymbol{S}_{\mathit{K}} \in \mathbb{R}^{B \times K \times HW}$ (where $HW=H \times W$) denote the indices and similarity scores of the Top-k memory items for each pixel. The similarity scores $\boldsymbol{S}_{\mathit{K}}$ are normalized using softmax to obtain the attention weights $\boldsymbol{W}_{\mathit{A}} \in \mathbb{R}^{B \times K \times HW}$, representing the degree of matching between each pixel and the Top-k memory items. Then, the corresponding memory features are retrieved from the memory bank using $\boldsymbol{Idx}_{\mathit{K}}$.  	The retrieved memory tensor is denoted as $\boldsymbol{Mem}_{\mathit{K}} \in \mathbb{R}^{B \times K \times HW \times D}$, which contains the features of the Top-k memory items associated with each pixel position. The weighted sum yields the compensation feature $\boldsymbol{F}_{\mathit{comp}} \in \mathbb{R}^{B \times HW \times D}$, and the final output $\boldsymbol{O}_{\mathit{S}}$ is obtained by reshaping. The weighted sum is computed as:
\begin{equation}
	\boldsymbol{F}_{\mathit{comp}}[b, hw,d] = \sum_{k=1}^{K} \boldsymbol{W}_{\mathit{A}}[b, k, hw] \cdot \boldsymbol{Mem}_{\mathit{K}}[b, k, hw,d]
\end{equation}

Each expert employs distinct convolutions to fuse the features from GPStream and SCStream, capturing specific semantic relations and enabling adaptive refinement.

\subsection{Nighttime Image Reflection Separation Dataset}
\begin{figure}[t]
	\centering
	\includegraphics[width=0.55\columnwidth]{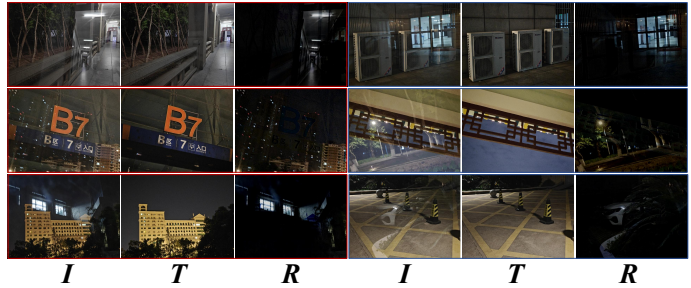} 
	\caption{Examples from the NightIRS dataset. $\boldsymbol{I}$, $\boldsymbol{T}$, and $\boldsymbol{R}$ denote the blended image, transmission layer, and reflection layer, respectively.}
	
	\label{fig_NightIRS}
\end{figure}

The Nighttime Image Reflection Separation (NightIRS) dataset contains 1,000 nighttime reflection image triplets. Each triplet consists of $\boldsymbol{I}$, $\boldsymbol{T}$, and $\boldsymbol{R}$, as shown in Figure~\ref{fig_NightIRS}. Reflection interference is introduced using glass and acrylic sheets of varying thicknesses. To ensure illumination diversity, the dataset is collected under various nighttime conditions, such as street lights, neon signs, illuminated buildings, and low-light natural environments.  To capture geometric variations of reflections, different camera-to-glass distances and viewing angles are considered.
The dataset also provides a high-resolution version (NightIRS-HR), offering scalable benchmarks for nighttime reflection separation.

\begin{table*}[t]
	\centering
	\setlength{\tabcolsep}{0pt}
	\resizebox{\textwidth}{!}{
		\begin{tabular}{c|ccc|ccc|ccc|ccc|ccc|ccc}
			\hline
			\multirow{2}{*}{\textbf{Methods}} 
			& \multicolumn{3}{c|}{\textbf{Nature (20)}} 
			& \multicolumn{3}{c|}{\textbf{Real (20)}} 
			& \multicolumn{3}{c|}{\textbf{Wild (55)}} 
			& \multicolumn{3}{c|}{\textbf{Postcard (199)}} 
			& \multicolumn{3}{c|}{\textbf{Solid (200)}} 
			& \multicolumn{3}{c}{\textbf{Average}} \\
			\cline{2-19}
			& PSNR↑ & SSIM↑ & LPIPS↓ 
			& PSNR↑ & SSIM↑ & LPIPS↓ 
			& PSNR↑ & SSIM↑ & LPIPS↓ 
			& PSNR↑ & SSIM↑ & LPIPS↓ 
			& PSNR↑ & SSIM↑ & LPIPS↓ 
			& PSNR↑ & SSIM↑ & LPIPS↓ \\
			\hline
			BDN (ECCV'18) & 18.83 & 0.737 & 0.242 & 18.68 & 0.728 & 0.284 & 22.02 & 0.822 & 0.181 & 20.54 & 0.857 & 0.177 & 22.68 & 0.856 & 0.125 & 20.55 & 0.800 & 0.202 \\
			ERRNet (CVPR'19) & 20.43 & 0.756 & 0.172 & 23.03 & 0.810 & 0.156 & 23.87 & 0.848 & 0.132 & 21.81 & 0.874 & 0.152 & 24.72 & 0.896 & 0.095 & 22.77 & 0.837 & 0.141 \\
			IBCLN (CVPR'20) & 23.78 & 0.784 & 0.145 & 21.59 & 0.764 & 0.210 & 24.46 & 0.885 & 0.134 & 22.95 & 0.875 & 0.155 & 24.74 & 0.893 & 0.097 & 23.50 & 0.840 & 0.148 \\
			LANet (ICCV'21) & 23.55 & 0.811 & 0.115 & 22.51 & 0.815 & 0.145 & 26.06 & 0.900 & 0.109 & 24.14 & 0.907 & 0.106 & 24.30 & 0.898 & 0.087 & 24.11 & 0.866 & 0.112 \\
			YTMT (NIPS'21) & 20.77 & 0.769 & 0.178 & 22.86 & 0.807 & 0.158 & 25.07 & 0.892 & 0.116 & 22.40 & 0.881 & 0.147 & 24.70 & 0.899 & 0.092 & 23.16 & 0.850 & 0.138 \\
			DMGN (TIP'21) & 20.63 & 0.764 & 0.167 & 20.28 & 0.763 & 0.215 & 21.34 & 0.774 & 0.152 & 22.65 & 0.879 & 0.151 & 23.27 & 0.872 & 0.102 & 21.63 & 0.810 & 0.157 \\
			HGNet (TNNLS'23) & 25.23 & 0.824 & 0.111 & 23.65 & 0.818 & 0.155 & 26.88 & 0.897 & 0.109 & 23.56 & 0.900 & 0.124 & 25.00 & 0.900 & 0.092 & 24.86 & 0.868 & 0.118 \\
			DSRNet (ICCV'23) & 21.62 & 0.781 & 0.149 & 23.41 & 0.805 & 0.147 & 24.35 & 0.893 & 0.117 & 24.66 & 0.911 & 0.111 & 26.10 & 0.914 & 0.071 & 24.03 & 0.861 & 0.119 \\
			RDRNet (CVPR'24) & 24.44 & 0.820 & \underline{0.107} & 21.29 & 0.769 & 0.190 & 26.48 & 0.905 & 0.101 & 23.65 & 0.891 & 0.146 & 25.93 & 0.912 & 0.080 & 24.36 & 0.860 & 0.125 \\
			DSIT (NIPS'24) & \underline{26.05} & \underline{0.830} & 0.128 & 24.34 & 0.823 & 0.136 & 27.55 & \textbf{0.920} & \textbf{0.081} & \textbf{26.01} & \textbf{0.921} & 0.103 & \underline{26.62} & \underline{0.922} & 0.075 & 26.11 & 0.883 & 0.105 \\
			
			RDNet (CVPR'25) & 25.77 & 0.828 &  0.108 & \textbf{25.13} & \textbf{0.838} & \textbf{0.117} & \underline{27.59} & 0.915 & 0.085 & \underline{25.95} & \textbf{0.921} & \textbf{0.088} & 26.59 & \underline{0.922} & \underline{0.069} & \underline{26.21} & \underline{0.885} & \underline{0.094} \\
			
			\rowcolor{gray!15}		DMDNet (Ours) & \textbf{26.68} & \textbf{0.838} & \textbf{0.097} & \underline{24.60} & \underline{0.836} & \underline{0.130} & \textbf{27.70} & \textbf{0.920} & \underline{0.083} & 25.32 & \textbf{0.921} & \underline{0.093} & \textbf{27.07} &\textbf{0.929} & \textbf{0.064} & \textbf{26.27} & \textbf{0.889} & \textbf{0.093} \\
			\hline
		\end{tabular}
	}
	\caption{
		Quantitative comparison of the transmission layer on public datasets. DMDNet achieves the best average performance.  
		\textbf{Bold} and \underline{underline} denote Top-1 and Top-2 results, respectively. 
		$\uparrow$ indicates higher is better, while $\downarrow$ indicates lower is better.
	}
	\label{tab:quantitative_comparison_T}
\end{table*}


\begin{table}[t]
	\centering
	\setlength{\tabcolsep}{13pt} 
	\resizebox{\columnwidth}{!}{
		\begin{tabular}{c|ccc|ccc|cc}
			\hline
			\multirow{2}{*}{\textbf{Methods}} 
			& \multicolumn{3}{c|}{\textbf{Transmission Layer}} 
			& \multicolumn{3}{c|}{\textbf{Reflection Layer}} 
			& \textbf{Param} & \textbf{FLOPs} \\
			\cline{2-7}
			& {PSNR↑} & {SSIM↑} & {LPIPS↓}
			& {PSNR↑} & {SSIM↑} & {LPIPS↓}
			& {(M)↓} & {(G)↓} \\
			\hline
			BDN (ECCV'18) & 20.52 & 0.680 & 0.293 & 8.79 & 0.082 & 0.843 & 75.16 & 12.70 \\
			ERRNet (CVPR'19) & 22.43 & 0.767 & 0.180    & N/A    & N/A    & N/A    & 18.95 & 116.72 \\
			IBCLN (CVPR'20) & 23.16 & 0.803 & 0.196 & 20.54 & 0.292 & 0.701 & 21.61 & 98.16 \\
			
			LANet (ICCV'21) & 23.68 & 0.817 & 0.171 & 21.61 & 0.280 & 0.472 & 10.93 & 83.81 \\
			YTMT (NIPS'21) & 23.03 & 0.799 & 0.186 & 24.96 & 0.500 & 0.503 & 76.90 & 110.98 \\
			DMGN (TIP'21) & 22.88 & 0.799 & 0.174 & 24.77 & 0.488 & 0.508 & 45.49 & 116.85 \\
			HGNet (TNNLS'23) & 23.60 & 0.817 & 0.170    & N/A    & N/A    & N/A    & 14.51  & 82.08 \\
			DSRNet (ICCV'23) & 23.39 & 0.813 & 0.175 & 24.80 & 0.404 & 0.499 & 124.6  & 90.21 \\
			RDRNet (CVPR'24) & 24.04 & 0.824 & 0.185    & N/A    & N/A    & N/A    & 29.09 & 5.14 \\
			DSIT (NIPS'24) & 24.61 & 0.827 & 0.168 & 27.18 & 0.569 & 0.372 & 131.76  & 74.18 \\
			RDNet (CVPR'25) & \underline{25.08} & \underline{0.831} & \underline{0.149} & \underline{27.93} & \textbf{0.636} & \underline{0.309} & 266.43    & 66.10 \\
			\rowcolor{gray!15}			DMDNet (Ours) & \textbf{25.24} & \textbf{0.832} & \textbf{0.144} & \textbf{28.37} & \underline{0.633} & \textbf{0.286} & 87.22 & 39.33 \\
			\hline
		\end{tabular}
	}
	\caption{Quantitative comparison with SOTAs on the NightIRS dataset. FLOPs for a 128×128 RGB image. }
	\label{tab:NightIRS_compared}
\end{table}
\section{Experiments}
\subsection{Implementation Details}
The channel dimensions are set as $C_1, C_2, C_3, C_4, C_5 = [48, 96, 192, 384, 768]$. In MECM, $N_{\mathit{Exp}} = 4$ and $N_{\mathit{Exp}}^{K} = 2$. We adopt a batch size of 1 and crop images into 352×352 patches. Random horizontal flipping is adopted for data augmentation during training. The model is optimized using the Adam optimizer \cite{kingma2014adam} with an initial learning rate of $10^{-4}$. We train for 60 epochs, and reduce the learning rate to $5 \times 10^{-5}$ and $10^{-5}$ at the 30th and 50th epochs, respectively. All experiments are conducted on a single NVIDIA RTX 4090 GPU. 
See supplementary material for more details.

\subsection{Dataset and Evaluation Metrics}
Following previous works \cite{zhao2025reversible,hu2024single,hu2023single,dong2021location}, we train our model on 7,643 image pairs from the PASCAL VOC dataset \cite{everingham2010pascal}, 200 image pairs from the Nature dataset \cite{li2020single}, and 89 image pairs from the Real dataset  \cite{zhang2018single}. The remaining images from the Nature and Real datasets, together with the Wild, Postcard, and Solid subsets from the SIR\textsuperscript{2} dataset \cite{wan2017benchmarking}, as well as the NightIRS dataset, are used for testing. To avoid GPU memory overflow, images from the Real dataset are resized by scaling the longer side to 420 pixels while preserving the original aspect ratio.

To ensure fairness, all output images are saved in lossless PNG format, and evaluation metrics are computed in the RGB color space, including PSNR \cite{huynh2008scope},  SSIM \cite{wang2004image}, and LPIPS \cite{zhang2018unreasonable}, which assess image quality from pixel-wise, structural, and perceptual perspectives, respectively.

\subsection{Performance Evaluation}
We compare our DMDNet with 11 methods, including  BDN \cite{yang2018seeing}, ERRNet \cite{wei2019single}, IBCLN \cite{li2020single}, LANet \cite{dong2021location}, YTMT \cite{hu2021trash}, DMGN \cite{feng2021deep}, HGNet \cite{zhu2023hue}, DSRNet \cite{hu2023single},  RDRNet \cite{zhu2024revisiting}, DSIT \cite{hu2024single}, and RDNet\cite{zhao2025reversible}. 
Table~\ref{tab:quantitative_comparison_T} presents a quantitative comparison on public datasets, which primarily consist of daytime scenes, demonstrating that DMDNet achieves the best average performance.
Table~\ref{tab:NightIRS_compared} presents a quantitative comparison on the NightIRS dataset. DMDNet attains the largest number of top-ranking metrics on both the transmission and reflection layers, demonstrating its adaptability to nighttime reflections, while maintaining a reasonable number of parameters and Floating-Point Operations (FLOPs).

\begin{figure*}[t]
	\centering
	\includegraphics[width=1.0\textwidth]{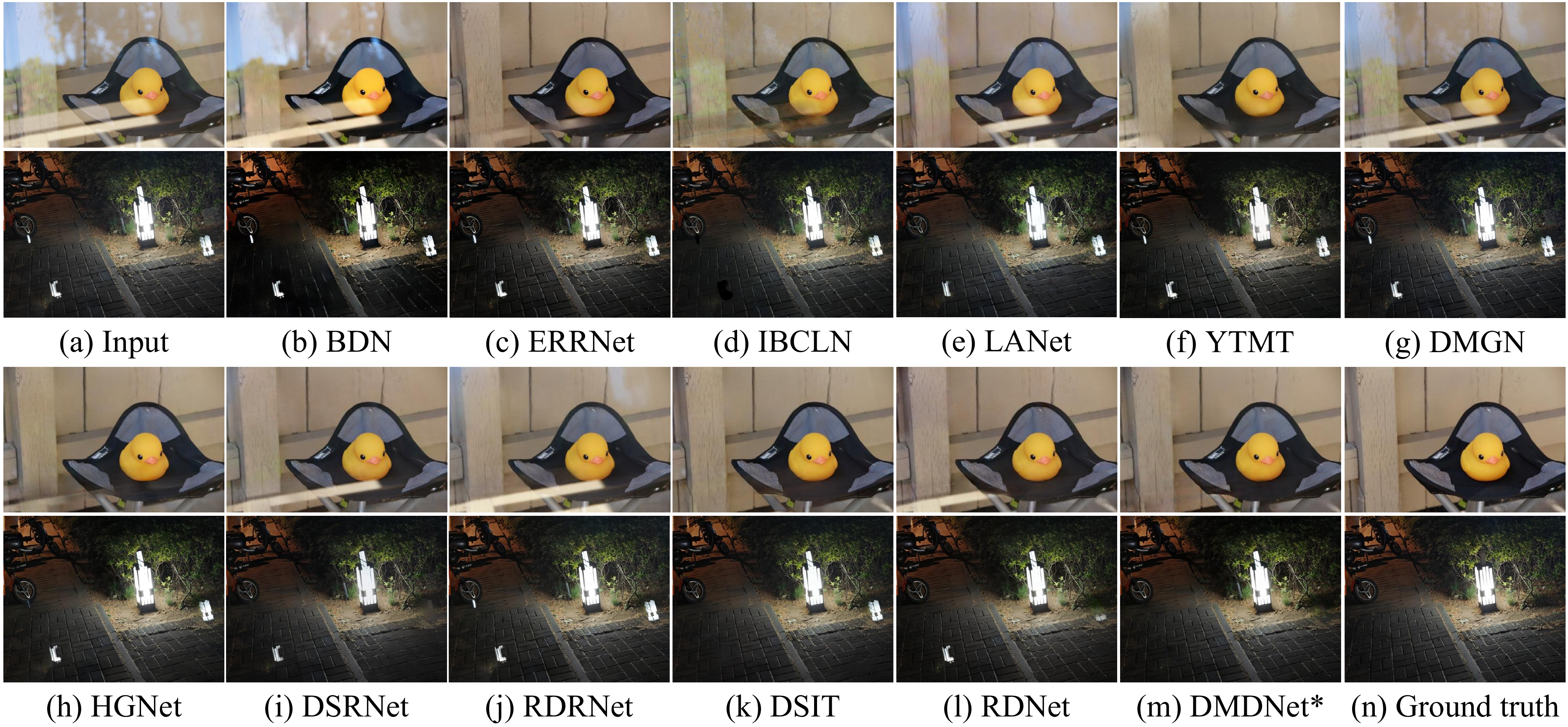}
	\caption{Qualitative comparison with SOTAs on the transmission layer. Our DMDNet removes reflections most effectively in both daytime and nighttime scenes. The nighttime image is taken from the NightIRS dataset.}
	\label{fig_compare_T}
\end{figure*}

Figure \ref{fig_compare_T} presents qualitative comparisons on the transmission layer. Our DMDNet achieves the most effective recovery, preserving structural details and suppressing residual reflections in daytime scenes. Even under nighttime conditions, where reflections closely resemble scene content, DMDNet effectively removes reflections. 

\subsection{Ablation Studies}

\subsubsection{DSMamba Visualization Analysis}
\begin{figure}[t]
	\centering
	\includegraphics[width=0.7\columnwidth]{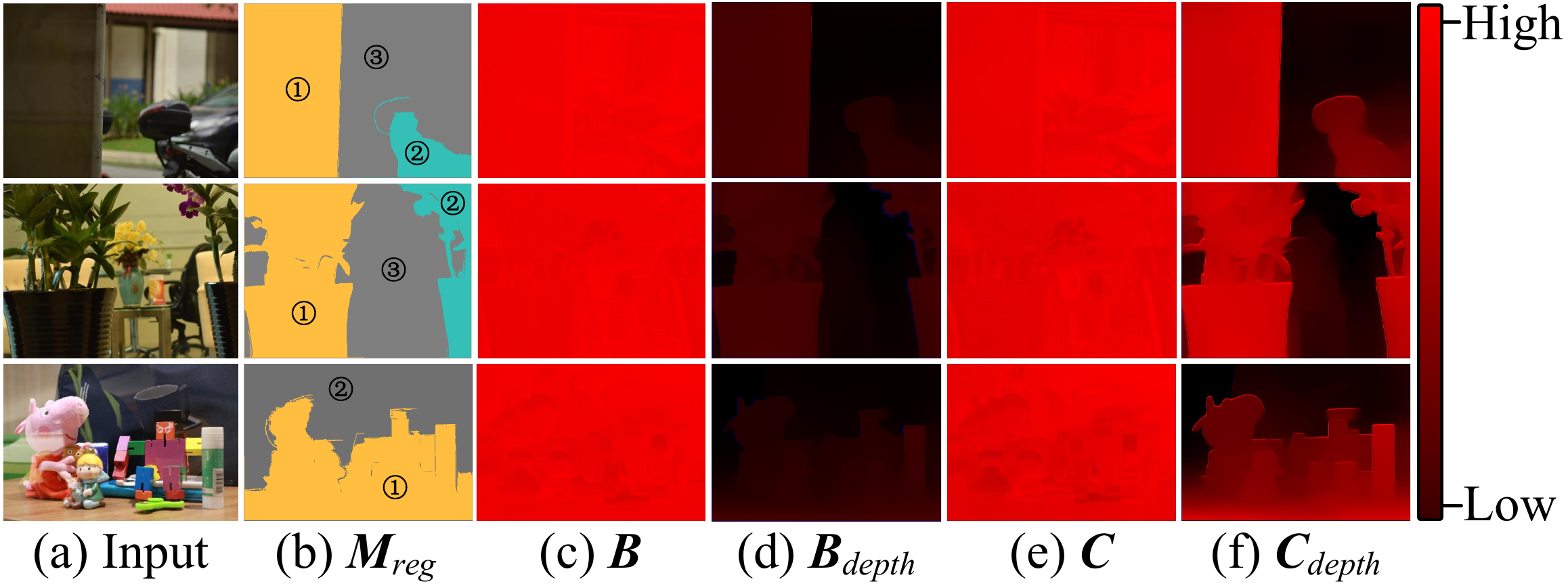} 
	\caption{DSMamba visualization.  $\boldsymbol{M}_{reg}$ shows the region-wise scanning order. (c)--(f) show the state-space matrices. $\boldsymbol{B}_{depth}$ and $\boldsymbol{C}_{depth}$ focus more on salient structures.}
	\label{fig_visual_depthMamba}
\end{figure}

Figure~\ref{fig_visual_depthMamba} (b) visualizes the scanning region map $\boldsymbol{M}_{reg}$ generated by DA-RScan. The partitioned regions align well with the structural layout. Figures~\ref{fig_visual_depthMamba}~(c)-(d) show that the original state-space matrices $\boldsymbol{B}$ and $\boldsymbol{C}$ exhibit a uniform distribution of activations, lacking discriminative focus. In contrast,  $\boldsymbol{B}_{depth}$ and $\boldsymbol{C}_{depth}$ amplify activations in salient structural regions while suppressing responses in ambiguous areas, improving the structural awareness of the state evolution. Notably, $\boldsymbol{B}_{depth}$ appears darker than $\boldsymbol{C}_{depth}$, as it more strictly regulates the influence of inputs on the state, resulting in generally lower activation values.

\subsubsection{Comparison with Mamba Variants}
For a fair comparison and to adapt these methods to reflection separation, we replace our DSMamba with MambaIR \cite{guo2024mambair}, VMambaIR \cite{shi2025vmambair}, and MambaIRv2 \cite{guo2025mambairv2} while keeping the training strategy identical. As shown in Table~\ref{tab:mamba_compare}, MambaIR and VMambaIR are constrained by fixed scanning orders, limiting their ability to disentangle overlapping layers. MambaIRv2’s attentive state-space design is easily disturbed by reflections with similar semantics to scene content. By contrast, our DSMamba outperforms these variants on both $\boldsymbol{T}$ and $\boldsymbol{R}$ restoration.

\begin{table}[t]
	\centering
	\setlength{\tabcolsep}{13pt}
	\resizebox{\columnwidth}{!}{
		\begin{tabular}{c|ccc|ccc|cc}
			\hline
			\multirow{2}{*}{\textbf{Methods}} 
			& \multicolumn{3}{c|}{\textbf{Transmission Layer}} 
			& \multicolumn{3}{c|}{\textbf{Reflection Layer}} 
			& \textbf{Param} & \textbf{FLOPs} \\
			\cline{2-7}
			& PSNR↑ & SSIM↑ & LPIPS↓
			& PSNR↑ & SSIM↑ & LPIPS↓
			& (M)↓ & (G)↓ \\
			\hline
			
			MambaIR   & 25.56 & 0.880 & 0.106 
			& \underline{22.09} & \underline{0.500} & 0.420 
			& 103.61 & 42.43 \\
			VMambaIR      & \underline{25.89} & \underline{0.884} & \underline{0.100} 
			& {22.06} & 0.490 & \underline{0.414} 
			& 83.76 & 38.05 \\
			MambaIRv2     & 24.84 & 0.868 & 0.118 
			& 21.66 & 0.490 & 0.445 
			& 88.38 & 40.38 \\
			\rowcolor{gray!15} DSMamba (Ours) & \textbf{26.27} & \textbf{0.889} & \textbf{0.093} & \textbf{22.31} & \textbf{0.522} & \textbf{0.403} & 87.22 & 39.33 \\
			\hline
		\end{tabular}
	}
	\caption{Comparison with Mamba variants on public datasets. FLOPs are calculated for a 128×128 RGB image.}
	\label{tab:mamba_compare}
\end{table}

\subsubsection{Ablation Study on DSMamba}
As shown in Table~\ref{tab:ablation_DSMamba}, the best performance is achieved when DA-RScan is used for $\boldsymbol{T}$ and DA-GScan for $\boldsymbol{R}$, outperforming the original four-directional scanning strategy in Vmamba. The results also demonstrate the superiority of DS-SSM over the original SSM, and validate the effectiveness of SPE.

\subsubsection{Ablation Study on MECM}
As shown in Table~\ref{tab:ablation_MECM}, both GPStream and SCStream are beneficial for performance. Furthermore, increasing the total number of memory experts $N_{\text{Exp}}$ offers more diverse feature priors, while selecting an appropriate number of top-k experts $N_{\text{Exp}}^K$ enables effective expert routing and reduces computational cost. The setting $N_{\mathit{Exp}}=4$, $N_{\mathit{Exp}}^K=2$ achieves a satisfactory balance.

\begin{table}[t]
	\centering
	\setlength{\tabcolsep}{13pt}
	\resizebox{\columnwidth}{!}{
		\begin{tabular}{cc|c|c|ccc|cc}
			\hline
			\multicolumn{2}{c|}{\textbf{Scanning Strategy}} 
			& \textbf{State-Space} 
			& \multirow{2}{*}{\textbf{SPE}} 
			& \multicolumn{3}{c|}{\textbf{Average}} 
			& \textbf{Param} & \textbf{FLOPs} \\
			\cline{1-2} \cline{5-7}
			$\boldsymbol{T}$ & $\boldsymbol{R}$ 
			& \textbf{Model} 
			&  
			& {PSNR~$\uparrow$} 
			& {SSIM~$\uparrow$} 
			& {LPIPS~$\downarrow$} 
			& {(M)↓} & {(G)↓} \\
			\hline
			\rowcolor{gray!15} DA-RScan & DA-GScan & DS-SSM & \checkmark & \textbf{26.27} & \textbf{0.889} & \textbf{0.093} & 87.22 & 39.33 \\
			DA-RScan    & DA-RScan   & DS-SSM   & \checkmark & 25.99 & 0.886 & 0.098 & 87.22 & 39.33 \\
			DA-GScan    & DA-GScan   & DS-SSM   & \checkmark & 25.87 & 0.886 & 0.100 & 87.22 & 39.33 \\
			DA-GScan    & DA-RScan   & DS-SSM   & \checkmark & \underline{26.09}    & \underline{0.887}    & \underline{0.096}    & 87.22 & 39.33 \\
			DA-RScan    & DA-GScan   & DS-SSM   & $\times$   & 25.66 & 0.882 & 0.105 & 87.22 & 39.33 \\
			DA-RScan    & DA-GScan   & Original & \checkmark & 25.78 & 0.884 & 0.098 & 83.29 & 38.55 \\
			Original & Original      & DS-SSM   & \checkmark & 25.69 & 0.884 & \underline{0.096} & 89.36 & 39.21 \\
			\hline
		\end{tabular}
	}
	\caption{Ablation study on DSMamba. Results are reported on the transmission layer of public datasets.}
	\label{tab:ablation_DSMamba}
\end{table}

\begin{table}[t]
	\centering
	\setlength{\tabcolsep}{15pt}
	\resizebox{\columnwidth}{!}{
		\begin{tabular}{cc|cc|ccc|cc}
			\hline
			\textbf{GP-} & \textbf{SC-} 
			& \multirow{2}{*}{$\boldsymbol{N_{\text{Exp}}^{K}}$} 
			& \multirow{2}{*}{$\boldsymbol{N_{\text{Exp}}}$} 
			& \multicolumn{3}{c|}{\textbf{Average}} 
			& \textbf{Param} & \textbf{FLOPs} \\
			\cline{5-7}
			\textbf{Stream} & \textbf{Stream} 
			& & 
			& {PSNR~$\uparrow$} 
			& {SSIM~$\uparrow$} 
			& {LPIPS~$\downarrow$} 
			& {(M)↓} & {(G)↓} \\
			\hline
			\rowcolor{gray!15} \checkmark & \checkmark & 2 & 4 & \textbf{26.27} & \textbf{0.889} & \textbf{0.093} & 87.22 & 39.33 \\
			$\times$   & $\times$   & 0 & 0 & 24.93 & 0.882 & 0.100 & 55.92 & 35.85 \\
			$\times$   & \checkmark & 2 & 4 & 25.91 & 0.884 & 0.100 & 80.98 & 37.66 \\
			\checkmark & $\times$   & 2 & 4 & \underline{26.07} & \underline{0.887} & \underline{0.096} & 80.98 & 37.67 \\
			\checkmark & \checkmark & 1 & 3 & 25.93 & 0.884 & 0.098 & 79.39 & 37.60 \\
			\hline
		\end{tabular}
	}
	\caption{Ablation study on MECM. Results are reported on the transmission layer of public datasets.}
	\label{tab:ablation_MECM}
\end{table}

\section{Conclusion}
We propose DMDNet to address the challenge of separating transmission and reflection layers when they exhibit similar contrast, especially in nighttime scenes. We present DSMamba, employing DAScan to prioritize structurally salient regions, and DS-SSM to enhance their influence on state evolution while suppressing the diffusion of interference. We introduce MECM, enabling experts to adaptively leverage cross-image knowledge to compensate for layer recovery. In addition, we construct the NightIRS dataset for evaluating nighttime reflection separation. Experimental results show that DMDNet outperforms SOTAs across all-day scenarios. 
One limitation is its reliance on supervised training data, and future work will explore unsupervised approaches.

\bibliographystyle{unsrt}

\bibliography{references}  






\clearpage
\appendix

\title{Supplementary Material: Depth-Synergized Mamba Meets Memory Experts for All-Day Image Reflection Separation}
\begin{center}
	{\Large\bfseries
		Supplementary Material: Depth-Synergized Mamba Meets Memory Experts for All-Day Image Reflection Separation
	}
	
	\vspace{1em}
	
\end{center}

This supplementary material provides additional details on the proposed DMDNet, 
including the design of loss functions, 
implementation settings, extended comparisons with state-of-the-art methods, 
supplemental ablation studies, 
statistical significance analysis, 
and a data appendix introducing the NightIRS dataset.

\section{Loss Functions}

The loss function consists of load loss $\mathcal{L}_{{load}}$, memory matching loss $\mathcal{L}_{{mem}}$, and appearance loss 	$\mathcal{L}_{{app}}$, which respectively balance expert usage, enhance memory matching, and ensure output fidelity.

The load loss uses the square of the coefficient of variation to balance the load of experts, and prevent certain experts from being overly relied upon. It is defined as:

\begin{equation}
	\mathcal{L}_{{load}} = \sum_{X \in \{T, R\}} \lambda^{{load}}_X \cdot \mathbb{E} \left[ \left( \frac{\sigma(W_X)}{\mu(W_X) + \epsilon} \right)^2 \right]
\end{equation}
Here, $W_X$ denotes the selection weights of each sample over all experts for layer $X$; 
$\sigma(\cdot)$ and $\mu(\cdot)$ represent the standard deviation and mean, respectively; 
$\epsilon$ is a small constant to avoid division by zero; 
$\mathbb{E}[\cdot]$ denotes the expectation operator; 
and $\lambda^{{load}}_X$ denotes the load loss weight for $\boldsymbol{T}$ and $\boldsymbol{R}$.

The memory matching loss encourages image features to be close to their most relevant memory items, while maintaining a clear margin from less relevant ones. To achieve this, $\mathcal{L}_{{mem}}$ consists of a triplet loss and a Mean Squared Error (MSE) loss, defined as follows:
\begin{align}
	\mathcal{L}_{{mem}} &= \sum_{X \in \{T, R\}} \bigg[
	\lambda_X^{{triplet}} \cdot 
\operatorname*{max}\!\left(
\|I_i - m_i^+\|_2^2 - \|I_i - m_i^-\|_2^2,\ 0
\right)
	+ \lambda_X^{{align}} \left\| I_i - m_i^+ \right\|_2^2
	\bigg]
\end{align}
Here, $I_i$ denotes the query feature from the image.
$m_i^+$ and $m_i^-$ represent the most and second most similar memory items to $I_i$, respectively. 
$\lambda_X^{{triplet}}$ and $\lambda_X^{{align}}$ are weighting coefficients that balance the contributions of the triplet term and the alignment term, respectively.

The appearance loss constrains the similarity between the restored images and the target images in both pixel and perceptual spaces. 
It consists of two components: a pixel-wise L1 loss and a perceptual loss based on VGG \cite{simonyan2014very} features:
\begin{equation}
	\begin{split}
		\mathcal{L}_{{app}} = 
		\lambda^{{L1}}_{T} \left\| \hat{T} - {T} \right\|_1 
		+ \lambda^{{L1}}_{R} \left\| \hat{{R}} - \mathcal{R} \right\|_1 
		+ \lambda^{{VGG}}_{T} \left\| {VGG}(\hat{T}) - {VGG}(T) \right\|_1
	\end{split}
\end{equation}
Here, $\hat{T}$ and $\hat{R}$ denote the restored transmission and reflection layers, respectively, 
while ${T}$ and ${R}$ represent the corresponding ground truth. $\lambda^{L1}_T$, $\lambda^{L1}_R$, and $\lambda^{VGG}_T$ are weighting coefficients that balance the contributions of the L1 and perceptual terms.

The overall loss function is defined as:
\begin{equation}
	\mathcal{L}_{{total}} = \mathcal{L}_{{load}} + \mathcal{L}_{{mem}} + \mathcal{L}_{{app}}
\end{equation}

\section{More Comparisons}

We further compare our DMDNet with 11 State-of-the-Art Methods (SOTAs), including BDN \cite{yang2018seeing}, ERRNet \cite{wei2019single}, IBCLN \cite{li2020single}, LANet \cite{dong2021location}, YTMT \cite{hu2021trash}, DMGN \cite{feng2021deep}, HGNet \cite{zhu2023hue}, DSRNet \cite{hu2023single}, RDRNet \cite{zhu2024revisiting}, DSIT \cite{hu2024single}, and RDNet \cite{zhao2025reversible}. 
On public datasets, including the Nature dataset \cite{li2020single}, the Real20 dataset \cite{zhang2018single}, and the Wild, Postcard, and Solid subsets from the SIR\textsuperscript{2} dataset \cite{wan2017benchmarking}, we evaluate the reflection layer recovery. As shown in  Table~\ref{tab:quantitative_comparison_reflect}, DMDNet achieves the largest number of Top-1 and Top-2 results, yielding the best average performance in reflection recovery. 

\begin{figure*}[t]
	\centering
	\includegraphics[width=0.85\textwidth]{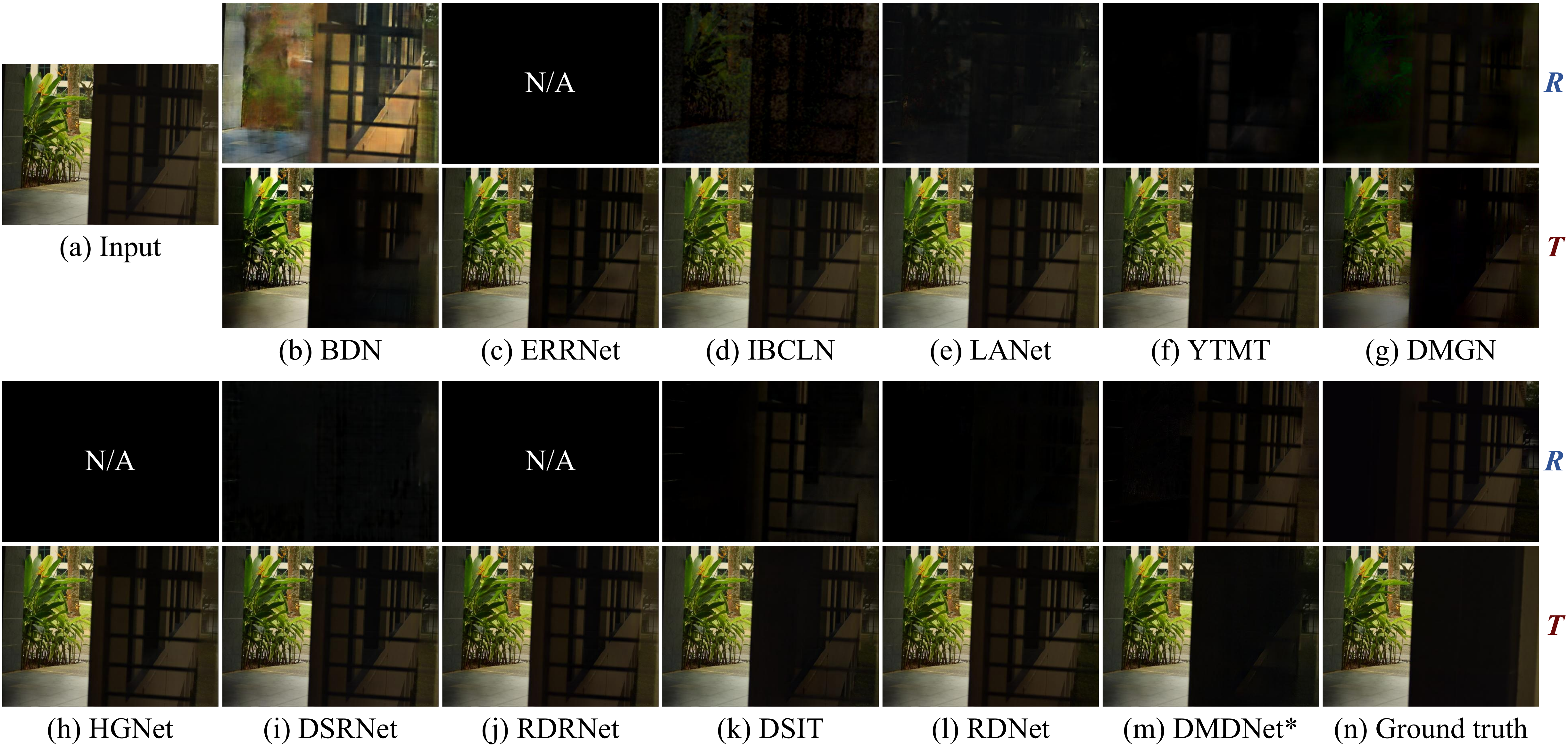}
	\caption{
		Qualitative comparison with SOTAs on daytime scenes. 
		Both transmission ($\boldsymbol{T}$) and reflection ($\boldsymbol{R}$) layers are shown for evaluation. DMDNet achieves improved $\boldsymbol{T}$ restoration with reduced reflection artifacts, while more faithfully recovering $\boldsymbol{R}$. 
		``N/A'' denotes absence of reflection layer output.
	}
	\label{wild_036}
\end{figure*}

\begin{figure*}[t]
	\centering
	\includegraphics[width=1.0\textwidth]{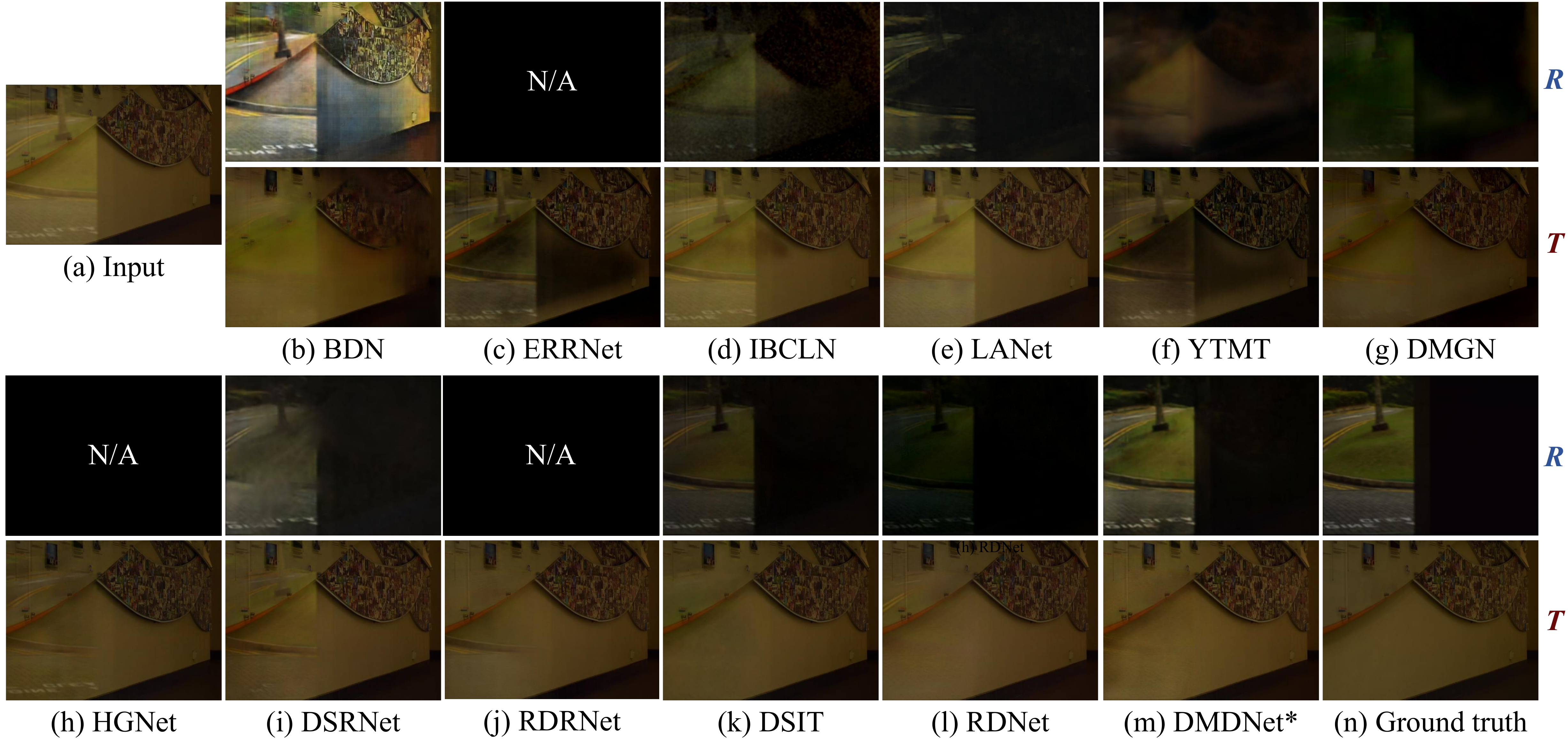}
	\caption{
		Qualitative comparison with SOTAs on indoor scenes. 
		DMDNet suppresses reflections more effectively, achieves clearer $\boldsymbol{T}$ restoration, and provides more faithful $\boldsymbol{R}$ recovery compared with other methods. 
	}
	
	\label{wild_054}
\end{figure*}

\begin{figure*}[t]
	\centering
	\includegraphics[width=1.0\textwidth]{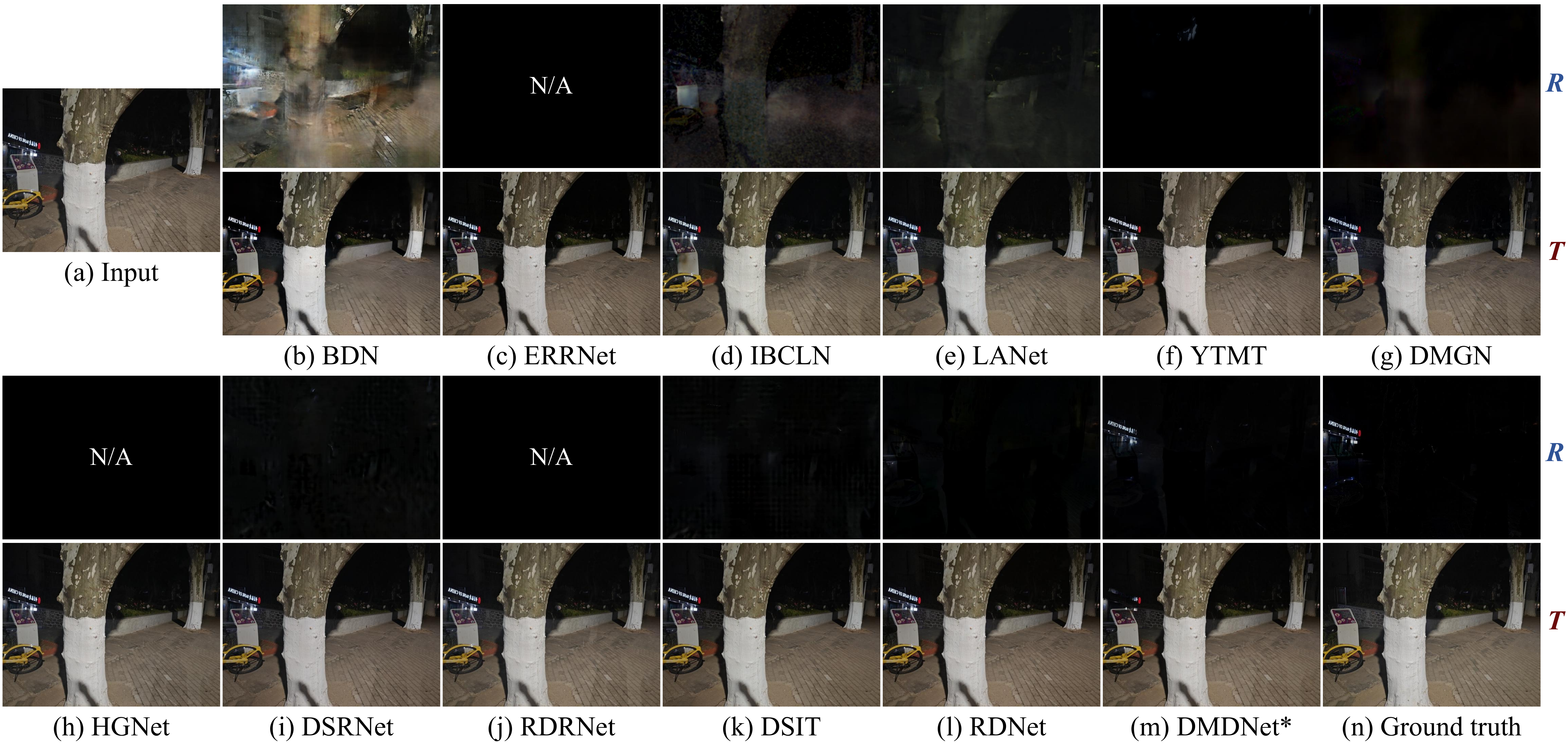}
	\caption{
		Qualitative comparison with SOTAs on nighttime roadside scenes. 
		DMDNet removes reflections more effectively, restores clearer $\boldsymbol{T}$ details under low-light conditions, and yields more faithful $\boldsymbol{R}$ recovery.  
	}
	\label{Night_IMG20250430224112_s5}
\end{figure*}

\begin{figure*}[t]
	\centering
	\includegraphics[width=1.0\textwidth]{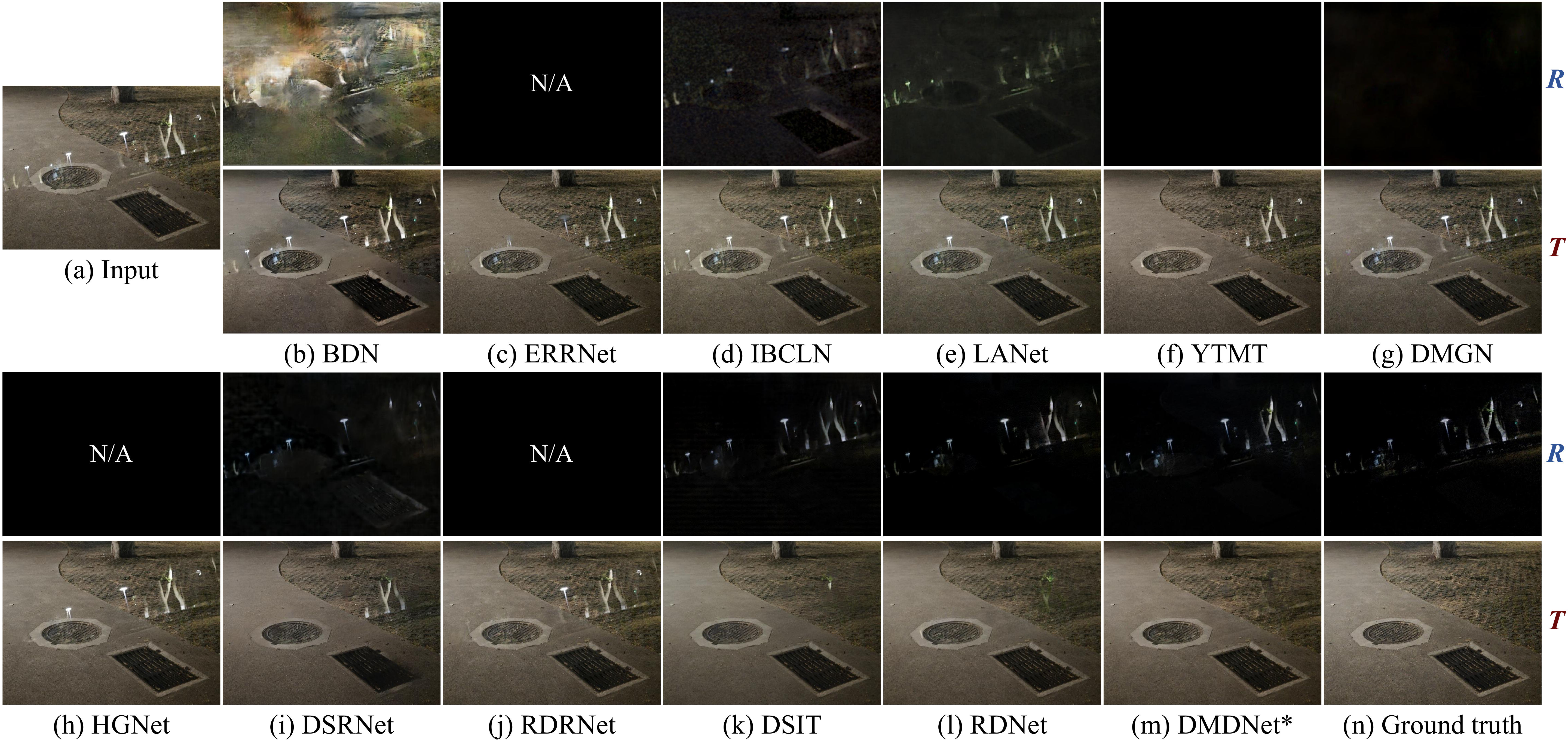}
	\caption{
		Qualitative comparison with SOTAs on nighttime ground scenes. 
		DMDNet suppresses reflections more effectively, restores clearer $\boldsymbol{T}$ details such as the pavement texture and manhole cover, and provides more faithful $\boldsymbol{R}$ recovery compared with other methods. 
	}
	\label{Night_IMG20250502225552_s1}
\end{figure*}

\begin{figure*}[t]
	\centering
	\includegraphics[width=1.0\textwidth]{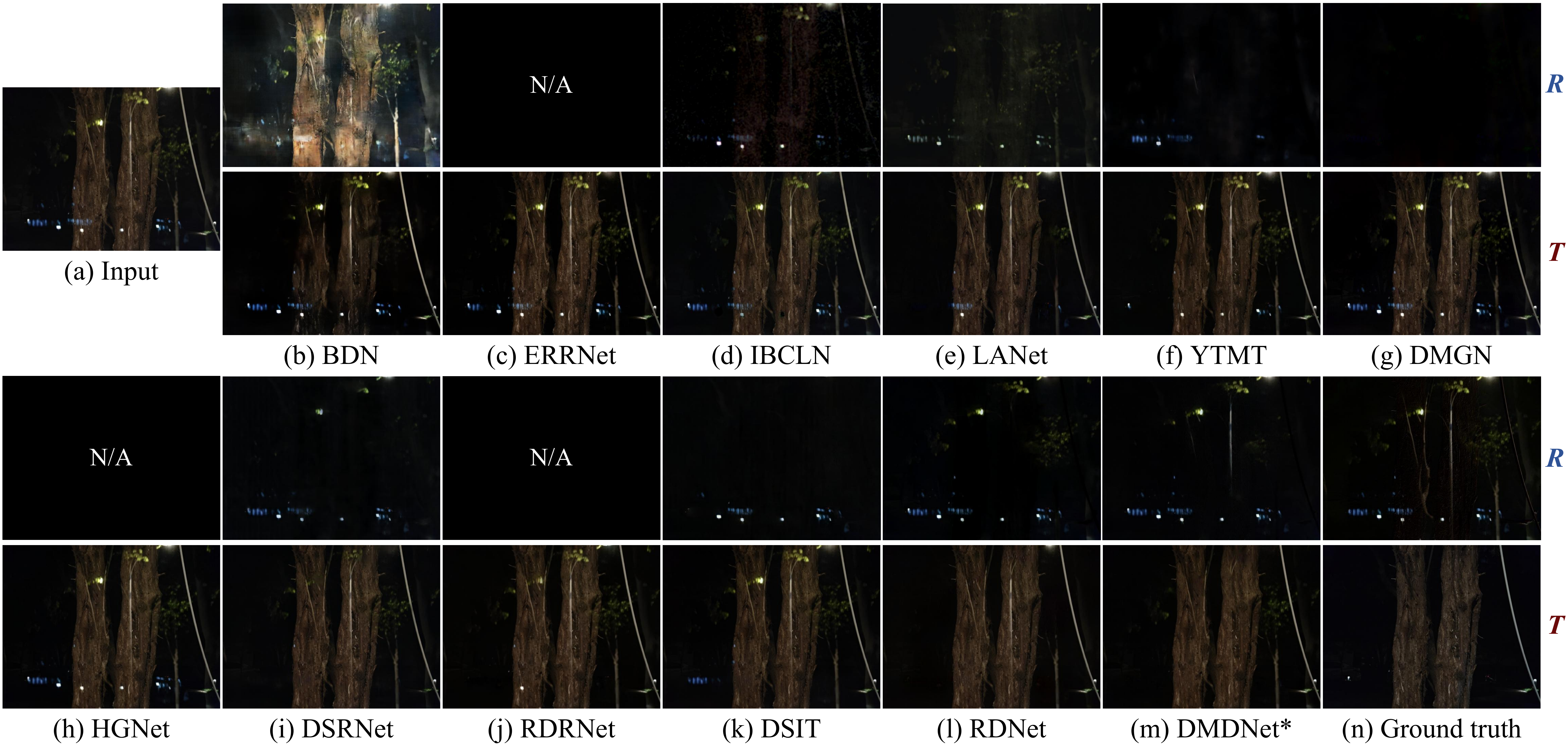}
	\caption{
		Qualitative comparison with SOTAs on nighttime natural scenes. 
		DMDNet better suppresses reflections from scattered lights, restores sharper $\boldsymbol{T}$ structures of trees, and more faithfully recovers $\boldsymbol{R}$ details.  
	}
	\label{Night_IMG20250430215301_s2}
\end{figure*}

We also compare DMDNet against the traditional methods L0-RS~\cite{arvanitopoulos2017single} and Fast-RS \cite{yang2019fast}. As shown in Table~\ref{tab:trad_methods}, our DMDNet outperforms both methods.
In addition, more qualitative comparisons of both the transmission layer ($\boldsymbol{T}$) and the reflection layer ($\boldsymbol{R}$) are presented in  Figures~\ref{wild_036}--\ref{Night_IMG20250430215301_s2}, further validating the effectiveness of DMDNet.

To assess human perceptual preference, we recruit 30 volunteers with normal visual function to perform subjective ranking of method outputs. Specifically, each participant evaluates 7 image groups, including 2 groups from Figure~\ref{fig_compare_T} and 5 groups from Figures~\ref{wild_036} to \ref{Night_IMG20250430215301_s2} , selecting the top 3 methods in each group. Our DMDNet receives 119 votes, exceeding the 91 votes of RDNet and the 87 votes of DSIT, indicating that DMDNet better aligns with human visual perception.

\section{Supplemental Ablation Studies}

\subsection{Ablation study on depth information}

To evaluate the influence of depth estimation accuracy, we replace the  depth model MiDaS v3.1 Next-ViT-L \cite{ranftl2022towards} with two lower-accuracy variants, v3.0 DPT-H and v2.1 DPT-Small. As shown in Table~\ref{tab:depth_ablation}, high-accuracy depth estimation leads to performance gains, while lower-accuracy depth incurs degradation. Even with lower-accuracy depth estimation, the model still surpasses the mainstream baseline RDRNet, demonstrating that our DSMamba can effectively extract useful cues even from coarse geometric structures. Further removing the depth prior from the network results in a noticeable drop in performance, highlighting the important role of depth information in the proposed method.

To investigate how depth information affects other methods, we introduce depth as additional prior to RDNet \cite{zhao2025reversible} and DSIT \cite{hu2024single}, concatenating it with the input and applying convolutional fusion. We load the original models as pretrained weights and train following their official settings. As shown in Table~\ref{tab:depth_ablation}, their performance deteriorates compared with the original versions, indicating that incorporation of depth disrupts their feature understanding. In contrast, DSMamba leverages depth information in an effective manner, fully exploiting the advantages of depth priors.

\begin{table*}[t]
	\centering
	\setlength{\tabcolsep}{0.8pt}
	\resizebox{\textwidth}{!}{
		\begin{tabular}{c|ccc|ccc|ccc|ccc|ccc|ccc}
			\hline
			\multirow{2}{*}{\textbf{Methods}} 
			& \multicolumn{3}{c|}{\textbf{Nature (20)}} 
			& \multicolumn{3}{c|}{\textbf{Real (20)}} 
			& \multicolumn{3}{c|}{\textbf{Wild (55)}} 
			& \multicolumn{3}{c|}{\textbf{Postcard (199)}} 
			& \multicolumn{3}{c|}{\textbf{Solid (200)}} 
			& \multicolumn{3}{c}{\textbf{Average}} \\
			\cline{2-19}
			& PSNR↑ & SSIM↑ & LPIPS↓ 
			& PSNR↑ & SSIM↑ & LPIPS↓ 
			& PSNR↑ & SSIM↑ & LPIPS↓ 
			& PSNR↑ & SSIM↑ & LPIPS↓ 
			& PSNR↑ & SSIM↑ & LPIPS↓ 
			& PSNR↑ & SSIM↑ & LPIPS↓ \\
			\hline
			BDN (ECCV'18) & 6.72 & 0.108 & 0.850 & 8.33 & 0.144 & 0.739 & 9.02 & 0.262 & 0.768 & 9.16 & 0.461 & 0.687 & 8.26 & 0.277 & 0.745 & 8.30 & 0.251 & 0.758 \\
			IBCLN (CVPR'20) & 16.96 & 0.298 & 0.757 & 18.60 & 0.325 & 0.666 & 19.54 & 0.482 & 0.690 & \underline{19.54} & 0.616 & 0.636 & 21.83 & 0.562 & 0.681 & 19.29 & 0.456 & 0.686 \\
			LANet (ICCV'21) & 18.79 & 0.335 & 0.604 & 19.57 & 0.383 & 0.481 & \underline{22.31} & \textbf{0.677} & 0.425 & \textbf{19.61} & \textbf{0.699} & \textbf{0.513} & \textbf{23.98} & \textbf{0.754} & 0.468 & 20.85 & \underline{0.570} & 0.498 \\
			YTMT (NIPS'21) & 21.23 & 0.339 & 0.647 & 22.51 & 0.453 & 0.531 & 20.15 & 0.187 & 0.429 & 11.92 & 0.153 & 0.811 & 18.77 & 0.061 & 0.553 & 18.92 & 0.238 & 0.594 \\
			DMGN (TIP'21) & 21.48 & 0.365 & 0.630 & 20.01 & 0.314 & 0.623 & 21.22 & 0.513 & 0.458 & 17.15 & 0.574 & \underline{0.607} & 20.57 & 0.399 & 0.522 & 20.08 & 0.433 & 0.568 \\
			DSRNet (ICCV'23) & 19.80 & 0.350 & 0.706 & 23.43 & 0.491 & 0.505 & 21.71 & \underline{0.643} & 0.455 & 18.47 & \underline{0.671} & 0.627 & \underline{23.16} & \underline{0.739} & 0.486 & 21.31 & \textbf{0.579} & 0.556 \\
			DSIT (NIPS'24) & 27.53 & 0.641 & 0.427 & 24.41 & 0.554 & 0.448 & \textbf{22.98} & 0.556 & \textbf{0.343} & 13.44 & 0.390 & 0.634 & 21.72 & 0.542 & \underline{0.422} & 22.02 & 0.537 & 0.455 \\
			RDNet (CVPR'25) & \underline{28.37} & \underline{0.657} & \underline{0.326} & \textbf{25.67} & \underline{0.601} & \textbf{0.309} & 21.44 & 0.326 & 0.379 & 14.56 & 0.418 & 0.661 & 20.22 & 0.268 & 0.454 & \underline{22.05} & 0.454 & \underline{0.426} \\
			\rowcolor{gray!15}	DMDNet (Ours) & \textbf{28.95} & \textbf{0.715} & \textbf{0.316} & \underline{25.53} & \textbf{0.642} & \underline{0.320} & 22.19 & 0.448 & \underline{0.357} & 13.51 & 0.341 & 0.609 & 21.38 & 0.462 & \textbf{0.414} & \textbf{22.31} & 0.522 & \textbf{0.403} \\
			\hline
		\end{tabular}
	}
	\caption{
		Quantitative comparison of the reflection layer on public datasets. DMDNet achieves the best average performance in recovering the reflection layer.  
		\textbf{Bold} and \underline{underline} denote Top-1 and Top-2 results, respectively. 
		$\uparrow$ indicates higher is better, while $\downarrow$ indicates lower is better.
	}
	\label{tab:quantitative_comparison_reflect}
\end{table*}

\begin{table}[t]
	\centering
	\setlength{\tabcolsep}{20pt}
	\resizebox{\columnwidth}{!}{
		\begin{tabular}{l|ccc|ccc}
			\hline
			\multirow{2}{*}{Method} 
			& \multicolumn{3}{c|}{Public Datasets} 
			& \multicolumn{3}{c}{NightIRS} \\
			\cline{2-7}
			& PSNR~$\uparrow$ & SSIM~$\uparrow$ & LPIPS~$\downarrow$
			& PSNR~$\uparrow$ & SSIM~$\uparrow$ & LPIPS~$\downarrow$ \\
			\hline
			L0-RS (CVPR'17)   
			& \underline{21.16} & \underline{0.794} & 0.206 
			& \underline{23.25} & \underline{0.789} & \underline{0.231} \\
			
			Fast-RS (CVPR'19) 
			& 20.55 & 0.792 & \underline{0.205} 
			& 22.70 & 0.777 & 0.244 \\
			
			\rowcolor{gray!15}{DMDNet (Ours)} 
			& \textbf{26.27} & \textbf{0.889} & \textbf{0.093}
			& \textbf{25.24} & \textbf{0.832} & \textbf{0.144} \\
			\hline
		\end{tabular}
	}
	\caption{Comparison with traditional methods on the transmission layer.}
	\label{tab:trad_methods}
\end{table}

\begin{table}[t]
	\centering
	\setlength{\tabcolsep}{5pt}
	\resizebox{\columnwidth}{!}{
		\begin{tabular}{c|c|ccc|ccc|ccc|ccc}
			\hline
			\multirow{3}{*}{Method} &
			\multirow{3}{*}{Depth} &
			\multicolumn{6}{c|}{Public Datasets} &
			\multicolumn{6}{c}{NightIRS} \\
			\cline{3-14}
			& &
			\multicolumn{3}{c|}{Transmission Layer} &
			\multicolumn{3}{c|}{Reflection Layer} &
			\multicolumn{3}{c|}{Transmission Layer} &
			\multicolumn{3}{c}{Reflection Layer} \\
			\cline{3-14}
			& &
			PSNR$\uparrow$ & SSIM$\uparrow$ & LPIPS$\downarrow$ &
			PSNR$\uparrow$ & SSIM$\uparrow$ & LPIPS$\downarrow$ &
			PSNR$\uparrow$ & SSIM$\uparrow$ & LPIPS$\downarrow$ &
			PSNR$\uparrow$ & SSIM$\uparrow$ & LPIPS$\downarrow$ \\
			\hline
			
			\rowcolor{gray!15}DMDNet & v3.1
			& \textbf{26.27} & \textbf{0.889} & \textbf{0.093}
			& \textbf{22.31} & \underline{0.522} & \textbf{0.403}
			& \textbf{25.24} & \textbf{0.832} & \textbf{0.144}
			& \textbf{28.37} & \underline{0.633} & \textbf{0.286} \\
			
			DMDNet & v3.0
			& \underline{25.53} & \underline{0.880} & \underline{0.104}
			& \underline{22.19} & \textbf{0.534} & \underline{0.414}
			& \underline{24.57} & \underline{0.827} & \underline{0.151}
			& \underline{27.31} & 0.616 & \underline{0.319} \\
			
			DMDNet & v2.1
			& 24.98 & 0.875 & 0.110
			& 21.87 & 0.513 & 0.423
			& 24.40 & 0.821 & 0.163
			& 28.32 & \textbf{0.681} & 0.318 \\
			
			DMDNet & -
			& 23.89 & 0.853 & 0.131
			& 21.24 & 0.505 & 0.466
			& 23.97 & 0.815 & 0.174
			& 26.39 & 0.564 & 0.339 \\
			
			DSIT & v3.1
			& 22.75 & 0.849 & 0.179
			& 21.81 & 0.547 & 0.510
			& 23.81 & 0.811 & 0.209
			& 26.71 & 0.518 & 0.413 \\
			
			RDNet & v3.1
			& 21.74 & 0.821 & 0.193
			& 19.69 & 0.306 & 0.495
			& 21.72 & 0.701 & 0.226
			& 26.27 & 0.548 & 0.371 \\
			\hline
		\end{tabular}
	}
	\caption{Ablation study on depth estimation quality and depth integration across methods.}
	\label{tab:depth_ablation}
	
\end{table}

\begin{table}[t] 
	\centering
	\setlength{\tabcolsep}{19pt}
	\resizebox{\columnwidth}{!}{
		\begin{tabular}{l|ccc|cc}
			\hline
			\multirow{2}{*}{\textbf{State-Space Model Modeling Strategies}} 
			& \multicolumn{3}{c|}{\textbf{Average}} 
			& \textbf{Param} & \textbf{FLOPs} \\
			\cline{2-4}
			& {PSNR~$\uparrow$} & {SSIM~$\uparrow$} & {LPIPS~$\downarrow$} 
			& {(M)} & {(G)} \\
			\hline
			\rowcolor{gray!15}
			\makecell[l]{
				$\boldsymbol{B}_{\mathit{aware}} = (1 - \gamma)\boldsymbol{B} + \gamma\boldsymbol{B}_{\mathit{depth}}$,\\
				$\boldsymbol{C}_{\mathit{aware}} = (1 - \gamma)\boldsymbol{C} + \gamma\boldsymbol{C}_{\mathit{depth}}$
			} 
			& \textbf{26.27} & \textbf{0.889} & \textbf{0.093} & 87.22 & 39.33 \\
			\hline

			$\boldsymbol{B}_{\mathit{aware}} = \boldsymbol{B},$ 
			& \multirow{2}{*}{25.78} & \multirow{2}{*}{0.884} & \multirow{2}{*}{{0.098}} & \multirow{2}{*}{83.29} & \multirow{2}{*}{38.55} \\
			$\boldsymbol{C}_{\mathit{aware}} = \boldsymbol{C}$ & & & & & \\
			\hline
			
			$\boldsymbol{B}_{\mathit{aware}} = \boldsymbol{B} + \boldsymbol{B}_{\mathit{depth}},$
			& \multirow{2}{*}{25.83} & \multirow{2}{*}{0.885} & \multirow{2}{*}{0.097} & \multirow{2}{*}{87.11} & \multirow{2}{*}{39.28} \\
			$\boldsymbol{C}_{\mathit{aware}} = \boldsymbol{C} + \boldsymbol{C}_{\mathit{depth}}$ & & & & & \\
			\hline
			
			$\boldsymbol{B}_{\mathit{aware}} = (1 - \gamma)\, \boldsymbol{B} + \gamma\, \boldsymbol{B}_{\mathit{depth}},$
			& \multirow{2}{*}{{\underline{26.24}}} & \multirow{2}{*}{0.886} & \multirow{2}{*}{\underline{0.095}} & \multirow{2}{*}{87.01} & \multirow{2}{*}{39.33} \\
			$\boldsymbol{C}_{\mathit{aware}} = \boldsymbol{C}$ & & & & & \\
			\hline
			
			$\boldsymbol{B}_{\mathit{aware}} = \boldsymbol{B},$ 
			& \multirow{2}{*}{{26.04}} & \multirow{2}{*}{\underline{0.887}} & \multirow{2}{*}{0.098} & \multirow{2}{*}{87.01} & \multirow{2}{*}{39.33} \\
			$\boldsymbol{C}_{\mathit{aware}} = (1 - \gamma)\, \boldsymbol{C} + \gamma\, \boldsymbol{C}_{\mathit{depth}}$ & & & & & \\
			\hline
		\end{tabular}
	}
	\caption{Ablation study on state-space modeling strategies. Results are reported on the transmission layer of public datasets. The proposed DS-SSM yields the highest overall performance.}
	\label{tab:ablation_statespace_dualrow}
\end{table}

\begin{table}[t]
	\centering
	\setlength{\tabcolsep}{14pt}
	\resizebox{\columnwidth}{!}{
		\begin{tabular}{ccccc|ccc|cc}
			\hline
			\multirow{2}{*}{$\boldsymbol{C_1}$} & \multirow{2}{*}{$\boldsymbol{C_2}$} & \multirow{2}{*}{$\boldsymbol{C_3}$} & \multirow{2}{*}{$\boldsymbol{C_4}$} & \multirow{2}{*}{$\boldsymbol{C_5}$}
			& \multicolumn{3}{c|}{\textbf{Average}} 
			& \textbf{Param} & \textbf{FLOPs} \\
			\cline{6-8}
			& & & & 
			& {PSNR~$\uparrow$} & {SSIM~$\uparrow$} & {LPIPS~$\downarrow$}
			& {(M)} & {(G)} \\
			\hline
			\rowcolor{gray!15}48  & 96   & 192  & 384  & 768   & \textbf{26.27} & \textbf{0.889} & \textbf{0.093} & 87.22  & 39.33 \\
			32  & 64   & 128  & 256  & 512   & 25.86 & 0.885 & \underline{0.096} & 39.36  & 22.98 \\
			64  & 128  & 256  & 512  & 1024  & \underline{26.06} & \underline{0.886} & 0.097 & 153.85 & 62.04 \\
			\hline
		\end{tabular}
	}
	\caption{Ablation study on the number of channels in DMDNet. Results are reported on the transmission layer of public datasets.}
	\label{tab:ablation_channels}
\end{table}

\begin{table}[t]
	\centering
	\setlength{\tabcolsep}{8pt}
	\resizebox{\columnwidth}{!}{
		\begin{tabular}{cc|cccc|ccc|ccc}
			\hline
			\multicolumn{2}{c|}{\textbf{Load Loss}} 
			& \multicolumn{4}{c|}{\textbf{Memory Matching Loss}} 
			& \multicolumn{3}{c|}{\textbf{Appearance Loss}} 
			& \multicolumn{3}{c}{\textbf{Average}} \\
			\hline
			$\lambda^{{load}}_T$ & $\lambda^{{load}}_R$ 
			& $\lambda^{{triplet}}_T$ & $\lambda^{{triplet}}_R$ 
			& $\lambda^{{align}}_T$ & $\lambda^{{align}}_R$ 
			& $\lambda^{L1}_T$ & $\lambda^{L1}_R$ & $\lambda^{{VGG}}_T$ 
			& {PSNR~$\uparrow$} & {SSIM~$\uparrow$} & {LPIPS~$\downarrow$} \\
			\hline
			\rowcolor{gray!15}	0.008 & 0.008 & 0.10 & 0.05 & 0.10 & 0.05 & 1 & 1 & 0.02 & \textbf{26.27} & \textbf{0.889} & \textbf{0.093} \\
			0     & 0     & 0.10 & 0.05 & 0.10 & 0.05 & 1 & 1 & 0.02 & 25.92 & \underline{0.885} & \underline{0.096} \\
			0.008 & 0.008 & 0    & 0    & 0    & 0    & 1 & 1 & 0.02 & 25.58 & 0.879 & 0.104 \\
			0     & 0     & 0    & 0    & 0    & 0    & 1 & 1 & 0.02 & 25.29 & 0.883 & 0.099 \\
			0     & 0     & 0    & 0    & 0    & 0    & 1 & 1 & 0    & 25.18 & 0.876 & 0.125 \\
			0.008 & 0.008 & 0.10 & 0.05 & 0.10 & 0.05 & 1 & 1 & 0    & 25.91 & 0.880 & 0.121 \\
			0.008 & 0.008 & 0.16 & 0.16 & 0.10 & 0.10 & 1 & 1 & 0.02 & 25.25 & 0.877 & 0.107    \\
			0.008 & 0.008 & 0.10 & 0.10 & 0.10 & 0.10 & 1 & 1 & 0.02 & \underline{26.03} & \underline{0.885}  & 0.098    \\
			\hline
		\end{tabular}
	}
	\caption{Ablation study on loss functions. Results are reported on the transmission layer of public datasets. }
	\label{tab:ablation_loss_weights}
\end{table}
\subsection{Ablation study on state-space modeling strategies.} 
We investigate different state-space modeling strategies by varying the formulations of $\boldsymbol{B}$ and $\boldsymbol{C}$, as summarized in Table~\ref{tab:ablation_statespace_dualrow}. 
When using the original matrices $\boldsymbol{B}$ and $\boldsymbol{C}$ without depth information integration, the model shows limited performance. 
Introducing depth-derived matrices $\boldsymbol{B}_{{depth}}$ and $\boldsymbol{C}_{{depth}}$ enhances the representational capacity, but directly adding them to the original matrices 
($\boldsymbol{B} = \boldsymbol{B} + \boldsymbol{B}_{{depth}}$, $\boldsymbol{C} = \boldsymbol{C} + \boldsymbol{C}_{{depth}}$) fails to yield optimal results.  
In contrast, adopting the Depth-Synergized State-Space Model (DS-SSM), i.e., $\boldsymbol{B}_{{aware}} = (1-\gamma)\boldsymbol{B} + \gamma \boldsymbol{B}_{{depth}}$ and $\boldsymbol{C}_{{aware}} = (1-\gamma)\boldsymbol{C} + \gamma \boldsymbol{C}_{{depth}}$, achieves the highest overall performance. 
These results demonstrate that the DS-SSM leads to improved layer separation quality while maintaining computational efficiency.

\subsection{Ablation study on the number of channels.}
We conduct an ablation study on different channel configurations in the main architecture of DMDNet, as shown in Table~\ref{tab:ablation_channels}. 
Using fewer channels greatly reduces the model size and computational cost but leads to a noticeable drop in performance. 
Increasing the number of channels generally improves performance, but excessively large channel sizes result in a sharp growth in parameters and Floating-Point Operations (FLOPs) without further gains, mainly due to the redundancy introduced by over-parameterization. 
The setting $(48, 96, 192, 384, 768)$ achieves the best trade-off between restoration quality and efficiency, confirming its suitability for the DMDNet architecture.

\subsection{Ablation study on loss functions.}
We further perform an ablation study on the loss functions of DMDNet, as summarized in Table~\ref{tab:ablation_loss_weights}. 
As shown in the first row of the table, our setting combines load loss, memory matching loss, and appearance loss with specific weight ratios, achieving the best overall performance.  
Removing certain loss terms or modifying their weights leads to performance degradation, demonstrating the effectiveness of our loss design.

\begin{table*}[t]
	\centering
	\setlength{\tabcolsep}{1pt}
	\resizebox{\textwidth}{!}{
		\begin{tabular}{c|cc|cc|cc|cc|cc|cc}
			\hline
			\multirow{3}{*}{\textbf{DMDNet vs.}} 
			& \multicolumn{6}{c|}{\textbf{Transmission Layer}} 
			& \multicolumn{6}{c}{\textbf{Reflection Layer}} \\
			\cline{2-13}
			& \multicolumn{2}{c|}{\textbf{PSNR}} 
			& \multicolumn{2}{c|}{\textbf{SSIM}} 
			& \multicolumn{2}{c|}{\textbf{LPIPS}}
			& \multicolumn{2}{c|}{\textbf{PSNR}} 
			& \multicolumn{2}{c|}{\textbf{SSIM}} 
			& \multicolumn{2}{c}{\textbf{LPIPS}} \\
			\cline{2-13}
			& \textbf{Statistic} & \textbf{P-value} 
			& \textbf{Statistic} & \textbf{P-value} 
			& \textbf{Statistic} & \textbf{P-value}
			& \textbf{Statistic} & \textbf{P-value} 
			& \textbf{Statistic} & \textbf{P-value} 
			& \textbf{Statistic} & \textbf{P-value} \\
			\hline
			BDN (ECCV'18) & \(1.4 \times 10^{4}\) & \(6.0 \times 10^{-234}\) & \(4.4 \times 10^{3}\) & \(6.0 \times 10^{-242}\) & \(1.1 \times 10^{3}\) & \(8.0 \times 10^{-245}\) & \(6.4 \times 10^{1}\) & \(1.1 \times 10^{-245}\) & \(4.3 \times 10^{4}\) & \(2.9 \times 10^{-209}\) & \(2.1 \times 10^{3}\) & \(7.0 \times 10^{-244}\) \\
			ERRNet (CVPR'19) & \(3.7 \times 10^{4}\) & \(1.0 \times 10^{-214}\) & \(1.6 \times 10^{4}\) & \(7.0 \times 10^{-232}\) & \(5.5 \times 10^{4}\) & \(2.7 \times 10^{-200}\) & N/A & N/A & N/A & N/A & N/A & N/A \\
			IBCLN (CVPR'20) & \(6.8 \times 10^{4}\) & \(8.0 \times 10^{-190}\) & \(7.8 \times 10^{4}\) & \(1.0 \times 10^{-182}\) & \(2.1 \times 10^{4}\) & \(2.0 \times 10^{-227}\) & \(1.8 \times 10^{5}\) & \(2.2 \times 10^{-112}\) & \(2.1 \times 10^{5}\) & \(2.8 \times 10^{-99}\) & \(8.4 \times 10^{3}\) & \(2.0 \times 10^{-238}\) \\
			LANet (ICCV'21) & \(1.5 \times 10^{5}\) & \(2.2 \times 10^{-135}\) & \(1.8 \times 10^{5}\) & \(2.2 \times 10^{-116}\) & \(1.3 \times 10^{5}\) & \(1.5 \times 10^{-146}\) & \(2.5 \times 10^{5}\) & \(2.2 \times 10^{-76}\) & \(3.2 \times 10^{5}\) & \(4.8 \times 10^{-46}\) & \(1.0 \times 10^{5}\) & \(7.0 \times 10^{-164}\) \\
			YTMT (NIPS'21) & \(6.4 \times 10^{4}\) & \(3.3 \times 10^{-193}\) & \(5.0 \times 10^{4}\) & \(6.0 \times 10^{-204}\) & \(4.8 \times 10^{4}\) & \(5.5 \times 10^{-206}\) & \(5.3 \times 10^{4}\) & \(4.9 \times 10^{-202}\) & \(1.3 \times 10^{5}\) & \(1.6 \times 10^{-145}\) & \(5.7 \times 10^{3}\) & \(8.0 \times 10^{-241}\) \\
			DMGN (TIP'21) & \(5.8 \times 10^{4}\) & \(1.2 \times 10^{-197}\) & \(4.5 \times 10^{4}\) & \(2.2 \times 10^{-208}\) & \(8.0 \times 10^{4}\) & \(7.5 \times 10^{-181}\) & \(2.3 \times 10^{5}\) & \(1.2 \times 10^{-87}\) & \(3.2 \times 10^{5}\) & \(6.1 \times 10^{-47}\) & \(2.7 \times 10^{4}\) & \(8.6 \times 10^{-223}\) \\
			HGNet (TNNLS'23) & \(1.2 \times 10^{5}\) & \(3.6 \times 10^{-152}\) & \(1.7 \times 10^{5}\) & \(1.5 \times 10^{-121}\) & \(7.6 \times 10^{4}\) & \(8.7 \times 10^{-184}\) & N/A & N/A & N/A & N/A & N/A & N/A \\
			DSRNet (ICCV'23) & \(1.6 \times 10^{5}\) & \(1.0 \times 10^{-124}\) & \(1.9 \times 10^{5}\) & \(1.3 \times 10^{-106}\) & \(1.6 \times 10^{5}\) & \(5.1 \times 10^{-128}\) & \(3.4 \times 10^{5}\) & \(5.0 \times 10^{-40}\) & \(4.1 \times 10^{5}\) & \(9.5 \times 10^{-20}\) & \(3.9 \times 10^{4}\) & \(6.0 \times 10^{-213}\) \\
			RDRNet (CVPR'24) & \(2.1 \times 10^{5}\) & \(1.7 \times 10^{-96}\) & \(2.7 \times 10^{5}\) & \(3.3 \times 10^{-66}\) & \(7.3 \times 10^{4}\) & \(4.1 \times 10^{-186}\) & N/A & N/A & N/A & N/A & N/A & N/A \\
			DSIT (NIPS'24) & \(4.1 \times 10^{5}\) & \(5.4 \times 10^{-18}\) & \(4.3 \times 10^{5}\) & \(5.5 \times 10^{-14}\) & \(2.1 \times 10^{5}\) & \(4.7 \times 10^{-99}\) & \(3.8 \times 10^{5}\) & \(4.0 \times 10^{-28}\) & \(4.0 \times 10^{5}\) & \(6.4 \times 10^{-21}\) & \(1.6 \times 10^{5}\) & \(3.5 \times 10^{-128}\) \\
			RDNet (CVPR'25) & \(5.3 \times 10^{5}\) & \(1.1 \times 10^{-1}\) & \(5.4 \times 10^{5}\) & \(2.4 \times 10^{-1}\) & \(5.4 \times 10^{5}\) & \(2.8 \times 10^{-1}\) & \(3.5 \times 10^{5}\) & \(2.2 \times 10^{-37}\) & \(3.6 \times 10^{5}\) & \(2.2 \times 10^{-33}\) & \(2.6 \times 10^{5}\) & \(4.9 \times 10^{-71}\) \\
			\hline
		\end{tabular}
	}
	\caption{Wilcoxon signed-rank test results of DMDNet against SOTAs, 
		summarized over both public datasets and the NightIRS dataset, 
		indicating that DMDNet achieves statistically significant improvements over most methods in both transmission and reflection layers.}

	\label{tab:statistics}
\end{table*}

\begin{figure*}[t]
	\centering
	\includegraphics[width=0.88\textwidth]{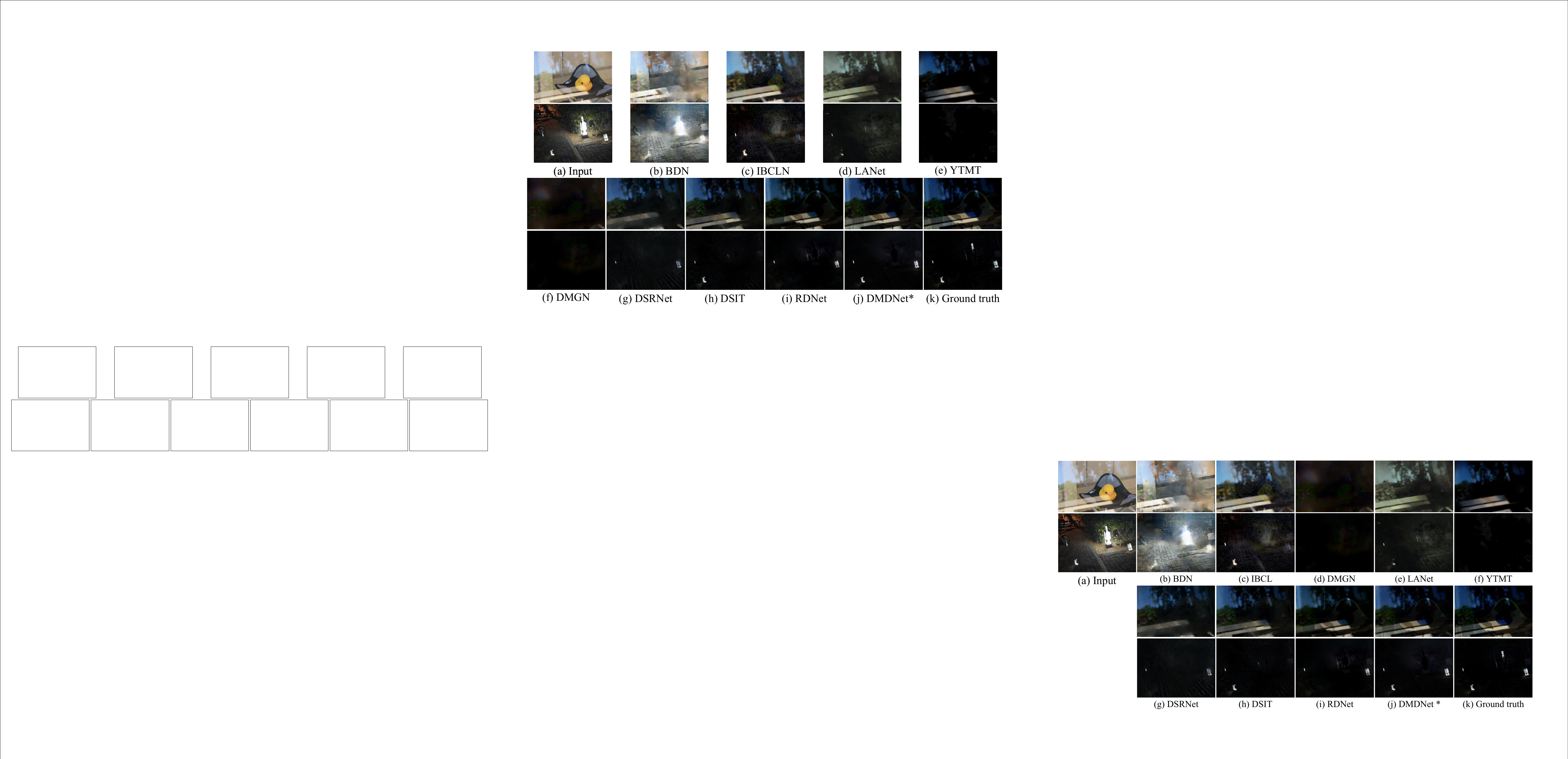}
	\caption{Qualitative comparison on the reflection layer corresponding to Figure~\ref{fig_compare_T}  in the main paper. 
		Compared with SOTAs, our DMDNet achieves more faithful reflection recovery.}
	
	\label{main-reflection}
\end{figure*}

\section{Further Implementation Details}
The loss weights are set as $\lambda^{{load}}_T = \lambda^{{load}}_R = 0.008$, $\lambda^{{triplet}}_T = \lambda^{{align}}_T = 0.1$, $\lambda^{{triplet}}_R = \lambda^{{align}}_R = 0.05$, $\lambda^{L\text{1}}_T = \lambda^{L\text{1}}_R = 1$, and $\lambda^{{VGG}}_T = 0.02$. 
For data synthesis, we adopt a widely used physical model~\cite{hu2023single}, formulated as
\begin{equation}
	\boldsymbol{I} = \alpha \boldsymbol{T} + \beta \boldsymbol{R} - \boldsymbol{T} \circ \boldsymbol{R},
\end{equation}
where $\boldsymbol{I}$ denotes the blended image, $\boldsymbol{T}$ the transmission layer, $\boldsymbol{R}$ the reflection layer, $\alpha$ and $\beta$ their respective blending coefficients, and $\circ$ the Hadamard product.

The model is trained on an Intel Xeon Platinum 8352V @ 2.10GHz, running Ubuntu 22.04.5 LTS, with Python 3.10.13, PyTorch 2.1.1, and CUDA 11.8, using a single NVIDIA RTX 4090 GPU. Each experiment is conducted twice, and the best performance is reported. When tested on an NVIDIA RTX 6000 Ada GPU, DMDNet takes 0.4 s and 4.2 GB of GPU memory to process a 512×512 RGB image, indicating a reasonable computational cost.

\section{Statistical Test}
We employ the Wilcoxon signed-rank test~\cite{wilcoxon1992individual} to assess the significance of the performance differences. 
As summarized in Table~\ref{tab:statistics}, the results show that DMDNet generally exhibits statistically significant advantages over existing methods in transmission and reflection layers.

\section{Data Appendix}
We introduce the Nighttime Image Reflection Separation (NightIRS) dataset to address the lack of benchmark data for reflection separation in nighttime scenes. Existing datasets (e.g., Nature \cite{li2020single}, Real \cite{zhang2018single}, SIR\textsuperscript{2} \cite{wan2017benchmarking}, RRW \cite{zhu2024revisiting}) predominantly contain daytime scenes with sufficient global illumination, which do not capture the challenges of nighttime conditions, where $\boldsymbol{T}$ and $\boldsymbol{R}$ often exhibit similar contrast and overlapping structures, owing to insufficient global illumination and scattered artificial lights.

The NightIRS dataset consists of 1,000 image triplets, each containing a blended image $\boldsymbol{I}$, a transmission layer $\boldsymbol{T}$, and a reflection layer $\boldsymbol{R}$. The images are captured using the Sony LYTIA‑T808, which provides high sensitivity in low-light conditions and HDR capability to faithfully record subtle nighttime details. Data collection is performed with the aid of a tripod for stability, and a wireless remote shutter to avoid vibration during capture. Acrylic and glass sheets of varying thicknesses (1~mm, 3~mm, 5~mm, and 8~mm) with a size of 700~mm $\times$ 500~mm are employed to introduce reflection interference. The dataset spans diverse nighttime conditions, including urban streets illuminated by artificial lights, indoor and outdoor reflection scenarios, and low-light natural environments, and its scale exceeds that of reflection removal datasets such as OpenRR-1k \cite{yang2025openrr} (83 samples) and RR4K \cite{chen2024real} (54 samples), providing a benchmark for advancing nighttime reflection separation research.

\begin{figure*}[t]
	\centering
	\includegraphics[width=1.0\textwidth]{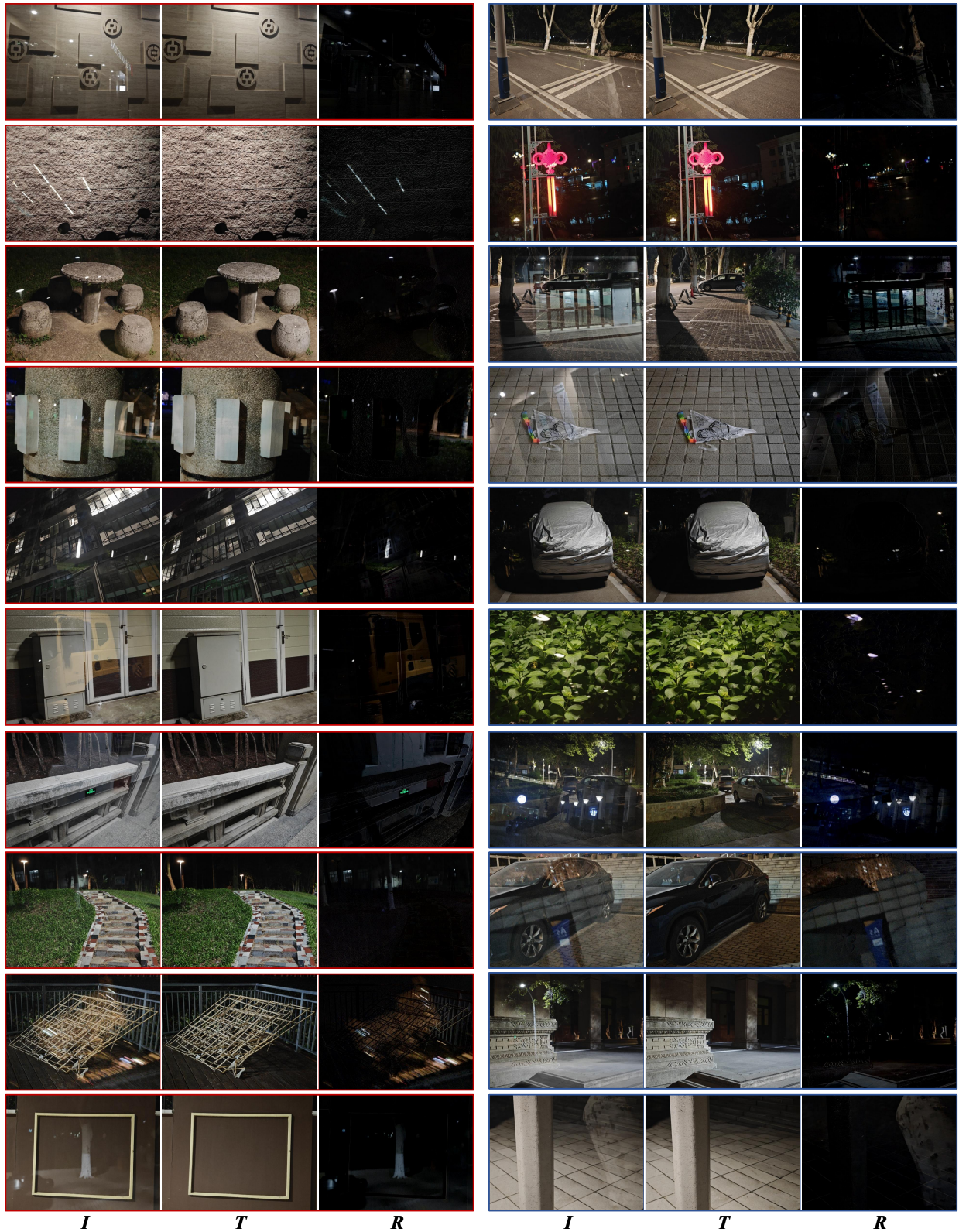}
	\caption{Examples from the NightIRS dataset. $\boldsymbol{I}$, $\boldsymbol{T}$, and $\boldsymbol{R}$ denote the blended image, transmission layer, and reflection layer, respectively.}
	\label{NightISR_sub}
\end{figure*}

Examples from NightIRS are shown in Figure~\ref{NightISR_sub}.  These examples cover diverse nighttime conditions of the dataset.

\bigskip

\end{document}